\documentclass[sigconf,natbib=true]{acmart}
\renewcommand\footnotetextcopyrightpermission[1]{} 

\AtBeginDocument{%
  \providecommand\BibTeX{{%
    \normalfont B\kern-0.5em{\scshape i\kern-0.25em b}\kern-0.8em\TeX}}}
\usepackage{microtype}

\usepackage[noend]{algpseudocode}
\usepackage{algorithm}
\usepackage{algorithmicx}
\usepackage{multirow}
\usepackage{microtype}
\usepackage{graphicx}
\usepackage[normalem]{ulem}
\usepackage{makecell}
\useunder{\uline}{\ul}{}
\usepackage{booktabs}
\usepackage{subfigure}
\usepackage{color}
\usepackage{amsmath}
\usepackage{bbding}
\usepackage{enumitem}
\usepackage{wrapfig}

\setcopyright{acmcopyright}
\copyrightyear{2018}
\acmYear{2018}
\acmDOI{XXXXXXX.XXXXXXX}

\acmConference[Conference acronym 'XX]{Make sure to enter the correct
  conference title from your rights confirmation emai}{June 03--05,
  2018}{Woodstock, NY}
\acmPrice{15.00}
\acmISBN{978-1-4503-XXXX-X/18/06}

\begin{document}

\title{MESED: A Multi-modal Entity Set Expansion Dataset with Fine-grained Semantic Classes and Hard Negative Entities}

\author{Yangning Li}
\authornote{These authors contributed equally to this research.}
\email{liyn20@mails.tsinghua.edu.cn}
\affiliation{%
  \institution{Tsinghua University}
  \city{PengCheng Laboratory}
  \state{}
  \country{}
}

\author{Tingwei Lu}
\authornotemark[1]
\email{ltw23@mails.tsinghua.edu.cn}
\affiliation{%
  \institution{Tsinghua University}
  \city{}
  \state{}
  \country{}
}

\author{Yinghui Li}
\authornotemark[1]
\email{liyinghu20@mails.tsinghua.edu.cn}
\affiliation{%
  \institution{Tsinghua University}
  \city{}
  \state{}
  \country{}
}

\author{Tianyu Yu}
\affiliation{%
  \institution{Tsinghua University}
  \city{}
  \state{}
  \country{}
}

\author{Shulin Huang}
\affiliation{%
  \institution{Tsinghua University}
  \city{}
  \state{}
  \country{}
}

\author{Hai-Tao Zheng}
\authornote{Corresponding author: zheng.haitao@sz.tsinghua.edu.cn.}
\affiliation{%
  \institution{Tsinghua University}
  \city{PengCheng Laboratory}
  \state{}
  \country{}
}

\author{Rui Zhang}
\affiliation{%
  \institution{Huawei Noah's Ark Lab}
  \city{}
  \state{}
  \country{}
}

\author{Yuan Jun}
\affiliation{%
  \institution{Huawei Noah's Ark Lab}
  \city{}
  \state{}
  \country{}
}

\renewcommand{\shortauthors}{Li, et al.}


\begin{abstract}

The Entity Set Expansion (ESE) task aims to expand a handful of seed entities with new entities belonging to the same semantic class. Conventional ESE methods are based on mono-modality (i.e., literal modality), which struggle to deal with complex entities in the real world such as: (1) Negative entities with fine-grained semantic differences. (2) Synonymous entities. (3) Polysemous entities. (4) Long-tailed entities. These challenges prompt us to propose Multi-modal Entity Set Expansion (MESE), where models integrate information from multiple modalities to represent entities. Intuitively, the benefits of multi-modal information for ESE are threefold: (1) Different modalities can provide complementary information. (2) Multi-modal information provides a unified signal via common visual properties for the same semantic class or entity. (3) Multi-modal information offers robust alignment signal for synonymous entities. To assess the performance of model in MESE and facilitate further research, we constructed the MESED dataset which is the first multi-modal dataset for ESE with large-scale and elaborate manual calibration. A powerful multi-modal model MultiExpan is proposed which is pre-trained on four multimodal pre-training tasks. The extensive experiments~\footnote{The benchmark and code will be public at \url{https://github.com/THUKElab/MESED}} and analyses on MESED demonstrate the high quality of the dataset and the effectiveness of our MultiExpan, as well as pointing the direction for future research.

\end{abstract}




\maketitle

\begin{figure}
    \centering
    \includegraphics[width=0.98\columnwidth]{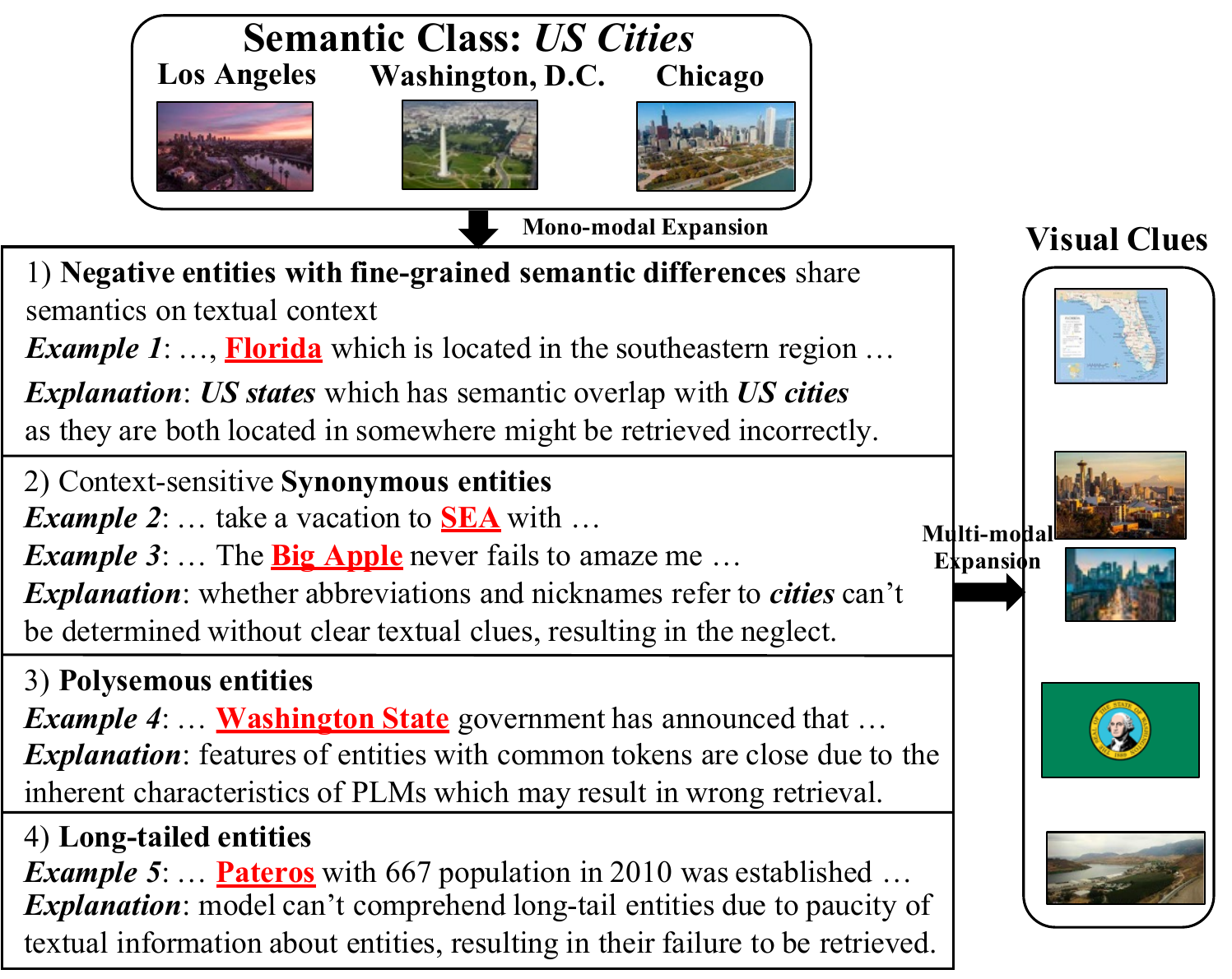}
    \caption{
    An example of tricky entities that a mono-modal ESE model cannot handle.
    }
    \label{fig:intro}
\end{figure}

\section{Introduction}
The Entity Set Expansion (ESE) task aims to expand a handful of seed entities with new entities belonging to the same semantic class based on the given candidate entity vocabulary and corpus~\cite{DBLP:journals/corr/abs-2207-08087, DBLP:journals/corr/abs-2304-03531}. For example, given \{\textit{Washington D.C.}, \textit{Chicago}, \textit{Los Angeles}\}, ESE tries to retrieve other entities with the target semantic class \texttt{US Cities}, such as \textit{New York}, \textit{NYC}, \textit{Boston}. ESE plays a significant role in knowledge mining and benefits a variety of downstream NLP and IR applications, such as web search\cite{10.1145/2835776.2835778}, taxonomy construction\cite{velardi-etal-2013-ontolearn, DBLP:journals/corr/abs-2211-10997}, and knowledge graphs~\cite{DBLP:journals/corr/abs-2302-08774, DBLP:journals/corr/abs-2304-04718, DBLP:journals/corr/abs-2211-04215}.

Conventional ESE methods are based on mono-modality (i.e., literal modality), which typically suffer from limited information and sparse representation. Taking expanding \texttt{US Cities} as an example, the mono-modal ESE methods struggle to deal with complex entities in the real world from the following perspectives:
\begin{itemize}[fullwidth,itemindent=1.5em]
    \item \textbf{Negative entities with fine-grained semantic differences} refer to entities that belong to the same coarse-grained semantic class as target class. These entities share semantics on textual context and are consequently challenging to be differentiated in detail. For instance, when expanding \texttt{US Cities}, it's inevitable to expand entities with the same parent class (i.e., \texttt{US Location}), such as \textit{Florida} and \textit{Texas} that are also located in the US.
    \item \textbf{Synonymous entities} mean entities have a variety of aliases. The ESE model can readily understand common aliases, while failing to comprehend these context-sensitive aliases \cite{henriksson2014synonym,schumacher2019learning, DBLP:journals/corr/abs-2306-12245} such as abbreviations and nicknames, since ascertaining the meaning of them necessitates explicit textual cues. For example, \textit{SEA} only means Seattle in certain contexts, potentially leading to the omission of its retrieval.
    \item \textbf{Polysemous entities}, which stand for possible ambiguity of a textual mention referring to multiple entities. Since pre-trained language models learn semantics through word co-occurrence \cite{kenton2019bert,lauscher2020specializing}, entities comprising the same tokens are inherently closer. For example, the L2 distance from \textit{Washington, D.C.} to \textit{Washington State} is instead smaller than the distance to many other cities like \textit{Austin} (8.89 vs. 10.02 we measured). As a result, entities merely with the same textual tokens may be wrongly retrieved.
    \item \textbf{Long-tailed entities} represent low-frequency entities in the corpus, such as obscure place names. Due to the inadequate textual description, the representation of these entities is frequently too sparse, posing a challenge to their retrieval.
\end{itemize}

The aforementioned situations lead to the advent of \textbf{Multi-modal Entity Set Expansion (MESE)}, where we integrate information from multiple modalities to represent entities and expand them to target semantic classes.

MESE can overcome the limitations of mono-modal approaches by leveraging multiple sources of information. The benefits of MESE include the following:
Firstly, multi-modal information can complement the information provided by texts (especially for short texts), thereby enhancing model to comprehensively understand entities. Secondly, multi-modal information can serve as a cohesive signal that unites semantic classes based on shared visual properties or characteristics. For instance, when dealing with \texttt{Comic Book Characters}, the background and style of images can serve as uniform features of the comic book characters, distinguishing them from hard negative semantic class \texttt{Movie Characters}. Third, multi-modal information can facilitate the resolution of polysemous entities and provide clues for the alignment of synonymous entities. In addition, we argue that multi-modal information is particularly beneficial to rarely used synonymous entities or long-tail entities, as entities of lower frequencies tend to be more concrete concepts with stable visual representations.

Regrettably, despite the availability of diverse multi-modal data types, there is currently no multi-modal dataset structured based on fine-grained semantic classes that can be utilized to evaluate the efficacy of MESE. To address this gap, we have constructed a large-scale, manually annotated MESE dataset called MESED, comprising 14,489 entities sourced from Wikipedia and 434,675 image-sentence pairs. To the best of our knowledge, MESED is the first multi-modal dataset for ESE with large-scale and elaborate manual calibration. MESED features several elements to accentuate the challenges of ESE. Firstly, we meticulously crafted a semantic class schema that consists of 26 coarse-grained and 70 fine-grained classes, with fine-grained classes that are mutually ambiguous (e.g., \textit{Chinese actors} versus \textit{US actors}) being assigned as hard negative classes for each other. Furthermore, synonymous and polysemous entities are added to amplify confusion between entities. Additionally, to gauge models' proficiency in comprehending sparse entities, uncommon semantic classes were deliberately included.

For benchmark settings, we designed three expansion settings based on MESED, including single text-based/visual-based expansion, and multi-modal expansion. Conventional text-based models, as well as emerging GPT-3.5, and various visual and multi-modal baseline models are evaluated. We also propose a powerful multi-modal model MultiExpan trained with four self-supervised multi-modal pre-training tasks that we designed, including masked entity prediction, contrastive learning, clustering learning, and momentum distillation.

To summarize, the main contributions are as follows:
\begin{itemize}[fullwidth,itemindent=1.5em]
\item We present a novel Multi-modal Entity Set Expansion (MESE) task, which expands entities in multiple modalities.
\item We first release a large-scale human-annotated MESE dataset called MESED, which is challenging as its fine-grained semantic classes and ambiguous candidate entities.
\item We provide strong multi-modal baseline models MultiExpan and explore diverse self-supervised pre-training objectives for representation learning of multi-modal entities.
\item Extensive experiments demonstrate the effectiveness of our MultiExpan compared to mono/multi-modal models. Detailed analyses are conducted to provide direction for future research.
\end{itemize}

\section{Related Work}
\subsection{Entity Set Expansion}
 Traditional ESE models \cite{kushilevitz2020a,mamou2019term,yu2019corpus,shen2017setexpan,rong2016egoset, huang2020guiding} usually comprehend entities based on their textual contexts. In recent years, the rapid development of Pre-trained Language Models (PLMs) \cite{kenton2019bert,lu2019vilbert,chen2020uniter,wang2021vlmo} has led to an increasing amount of work \cite{zhang2020empower,li2022contrastive} based on PLMs. \citet{li2022contrastive} proposed the ProbExpan, which employs contrastive learning among entities to obtain clearer semantic boundaries. The core challenge of the ESE task is to distinguish negative class entities from target entities in different situations; however, due to the inherent ambiguity of textual modality, as shown in Figure \ref{fig:intro}, the mono-modal ESE model is difficult to solve for complex entities, prompting us to propose the MESE task and attempt to construct a multi-modal ESE benchmark. 

For the ESE task, several mono-modal datasets are available. For example, Wiki \cite{shen2017setexpan} and APR \cite{shen2017setexpan} were constructed from Wikipedia and Reuters, respectively. Some previous work also used classic Named Entity Recognition datasets such as CoNLL \cite{zupon2019lightly} and OntoNotes \cite{sang2003introduction} as benchmarks for evaluation. However, these datasets do not highlight the previously mentioned challenges of ESE tasks. In terms of semantic class granularity, the Wiki with the most semantic classes has only eight semantic classes, and, more importantly, the semantic classes are so different from each other (e.g., \texttt{Countries} and \texttt{Diseases}) that there are not many hard negative class entities among the candidate entities, which significantly reduces the difficulty of the ESE task and causes assessment bias. 

The Appendix also covers multi-modal entity datasets relevant to ESE, such as those for multi-modal named entity recognition \cite{ding2021few} and entity linking \cite{wang2022wikidiverse,sun2022visual}, and explains why these datasets were not used to create MESE datasets.

\subsection{Vision-language Pre-trained Model}
Most existing works on vision-language representation learning fall into two categories. The first category \cite{su2020vl, lu2019vilbert,lu2020multi,tan2019lxmert,chen2020uniter,li2019visualbert} strives to model the deep interactions between image and text features with transformer-based multimodal encoders. Methods in this category achieve state-of-the-art results on multi-modal understanding tasks with image-text pairs as input (e.g., NLVR2 \cite{suhr2019a}, VQA \cite{antol2015vqa}). The second category \cite{radford2021learning,li2021align,faghri2018vse,DBLP:journals/patterns/LiuLTLZ22} focuses on learning separate encoders for each modality. The recent CLIP\cite{radford2021learning} and ALBEF \cite{li2021align} models are representative of this branch. These methods are well-suited for cross-modal retrieval because they encode each modality separately and are highly efficient. We believe that the first category of method is better aligned with the requirements of the ESE task, as it permits deep interactions and reasoning with multimodal contexts, leading to a more comprehensive comprehension of entities.

In addition, the impressive zero-shot performance of the emerging GPT-4 on a variety of multi-modal tasks made us impossible to ignore it~\cite{DBLP:journals/csur/DongLGCLSY23, DBLP:journals/corr/abs-2307-09007}. Unlike the two encoder-only models above, the decoder-only GPT-4 is a generative model \cite{brown2020language, ma-etal-2022-linguistic} that generates answers via beam search. Despite the excellent performance achieved, GPT-4 is unable to explicitly use the entire large vocabulary as input to constrain the output for ESE task, so the generated entities are not fully controlled and maybe out of the vocabulary.


\section{Task Formulation}
\textbf{Definition 1 \textit{Multi-modal Entity Set Expansion (MESE).}} The inputs of MESE are a small set $S=\{e_1,e_2,...,e_k\}$ that contains several seed entities describing a certain semantic class and a vocabulary $V$ of candidate entities. Besides, a corpus $D$ containing the multi-modal contexts $\{e_i,(t_1^i,v_1^i),...,(t_n^i,v_n^i)\}$ for each entity $e_i$ is given, in which $t_n^i$ is a sentence comprising $e_i$ and $(t_n^i,v_n^i)$ forms an image-sentence pair. It is of note that arbitrary modality may be lacking in a given context.

\section{Dataset Construction}
In this section, we demonstrate the MESED construction procedure. Several factors, including the coverage and ambiguity of semantic classes, as well as the relevance between images and entities are considered to ensure the quality of MESED.

\subsection{Data Collection}
\label{sec:data_collect}
There are two ways to construct a multi-modal ESE dataset. The first straightforward approach is to first collect the image-sentence pairs and label the entities in the sentences. Then, for each semantic class, human annotators traverse the entire large-scale entity vocabulary once to filter out the corresponding entities. Although plenty of public datasets are available with massive image-sentence pairs, the labour cost of such a bottom-up manner is prohibitive, and the construction process is not generalizable or scalable. Furthermore, the distribution of semantic class is contingent upon the specific dataset, rendering its regulation difficult. We therefore adopt the more general top-down approach to constructing MESED. That is, the semantic classes and the corresponding entities are constructed first, and then the text and visual contexts corresponding to the entities are collected in turn.

\noindent\textbf{Step 1. Semantic Classes and Entities Collection} 
Wikipedia has compiled a vast list of entities corresponding to semantic classes\footnote{https://en.wikipedia.org/wiki/List\_of\_lists\_of\_lists}, which are organized in a hierarchical structure. We pick a selection of semantic classes with certain principles (discussed in Section \ref{sec:analysis_of_mesed}) and crawl the corresponding entities. In addition, numerous entities randomly sampled from Wikipedia pages are appended to the entity vocabulary as negative entities. Further, polysemous and synonymous entities are also added to the vocabulary as hard negative entities and hard positive entities, respectively, which will also be discussed in Section \ref{sec:analysis_of_mesed}.

\noindent\textbf{Step 2. Entity-Labeled Sentences Collection} We crawl Wikipedia articles containing abundant entity mentions with human-annotated hyperlinks\footnote{Such as https://en.wikipedia.org/wiki/Universe} that uniquely identify an entity. Since the entities crawled in Step 1 contain hyperlinks, we can utilize these hyperlinks to associate the entities with the respective sentences and convey the textual information to the entities.

\noindent\textbf{Step 3. Related Images Collection} In this step, images corresponding to the entities or sentences are acquired through the Google Image search engine. To remove the distraction of extraneous content in the sentence, keywords in the sentence are extracted with KeyBERT \cite{grootendorst2020keybert}. We stitch them together with the entity name and semantic class as the search query, and obtain the top 10 images of the search results.

\noindent\textbf{Step 4. Images Re-ranking} One of the 10 images needs to be selected as the visual information of the entity. An ideal image should reflect the content of the sentence and contain the entity simultaneously. With both aspects in mind, a simple but effective image re-ranking algorithm was devised to select the most appropriate image $v_i$ for sentence $t_i$ 
and entity $e$: 
\begin{equation}
\begin{split}
    score(v_i, t_i, e)=\alpha \operatorname{CLIP-IMG}(v_i)\odot \operatorname{CLIP-TEXT}(t_i) \\ +(1-\alpha)\mathop{\max}_{o_i^j\in \operatorname{Obj}(v_i)}(\operatorname{cos\_sim}(o_i^j, \operatorname{Img}(e)))
\end{split}
\end{equation}
The first term measures the relevance of image $v_i$ and sentence $t_i$, which is what CLIP excels at. The second leveraged FasterRCNN \cite{ren2015faster} to detect objects $\operatorname{Obj}(v_i)$ in image and calculate their similarity to typical image $\operatorname{Img}(e)$ of entity in the Wikipedia Infobox. The second determines whether the entity appears in the image or not. We take the image with the highest score as the one corresponding to the sentence $t_i$ and entity $e$ and leave the exploitation of multiple images for future research.

Note that the above steps are automatic and labour-free.

\subsection{Calibration and Annotation}
The dataset automatically generated after the above steps is inevitably noisy. Especially in Steps 3 and 4, A mismatch between images and sentences  may exist. To improve the quality of images while verifying the effectiveness of the re-ranking algorithm, we hired human annotators who were required to evaluate the relevance of images to sentences and entities, categorized into three categories: relevant to both (R/T E\&S), relevant to only the sentence (R/T S), and irrelevant to both (IR). For images that are irrelevant to both after re-ranking, the annotators need to select a new image. 

From Table \ref{tab:data1}, we observe that the re-ranking algorithm significantly improves the relevance of images to both text and entities, compared to using the Top 1 image returned by the search engine directly. The inter-annotator agreement measured by Fleiss’s Kappa \cite{fleiss1971measuring} all exceeded 0.8, demonstrating the reliability of the annotation results. The strategy using the Top 1 image has the highest image diversity (measured by the inverse of the average cosine similarity of image embeddings) due to the introduction of substantial irrelevant images. Whereas the first term of the re-ranking algorithm guarantees the relevance of images and sentences while also avoiding a singular selection of typical images of the entity, potentially ensuring that there is no significant decrease in image diversity.

\begin{table}[htb]
\centering
\caption{Relevance of images between entities and sentences when using different strategies to process images.}
\scalebox{0.9}{
\begin{tabular}{cccccc}
\toprule
 Strategy         & R/T E\&S (\%) & R/T S (\%) & IR (\%) & Kappa & Diversity \\ \midrule
Top 1       & 52.7  & 14.8  & 32.5  & 0.842 & 1.813    \\ \midrule
Re-ranking       & 78.1  & 15.2  & 6.7  & 0.862 &  1.792 \\ \midrule
Annotation      & 80.8  & 19.2  & 0  & 0.858  & 1.798  \\ \bottomrule
\end{tabular}}
\label{tab:data1}
\end{table}

\subsection{Analysis of MESED}
\label{sec:analysis_of_mesed}
\noindent\textbf{Statistics of MESED}

MESED is the first multi-modal ESE dataset with meticulous manual calibration. It consists of 14,489 entities collected from Wikipedia, and 434,675 image-sentence pairs. The 70 fine-grained semantic classes in MESED contain an average of 82 entities with a minimum of 23 and a maximum of 362. Each fine-grained class contains 5 queries with three seed entities and 5 queries with five seed entities. MESED may not feature the largest total number of candidate entities, but we believe that the number of entities is not a key factor in measuring the quality of a dataset. Most candidate entities in previous datasets are randomly selected negative entities, which are significantly different from the target entities and do not enhance the challenge of the dataset.

\begin{table}[ht]
\small
\centering
\caption{Comparison of ESE datasets.}
\scalebox{0.9}{
\begin{tabular}{l|cccccc}
\toprule
 & \textbf{Wiki} & \textbf{APR} & \textbf{CoNLL} & \textbf{ONs} &\textbf{MESED} \\ \midrule
\# Semantic Classes & 8 & 3 & 4 & 8 &70 \\
Semantic granularity & Coarse & Coarse & Coarse & Coarse &Fine \\
\# Queries per Class & 5 & 5 & 1 & 1  & 10 \\
\# Seed Entities per Query & 3 & 3 & 10 & 10 &3/5\\
\# Candidate Entities & 33K & 76K & 6K & 20K &14K \\
\# Sentences of Corpus & 973K & 1043K & 21K & 144K &434K\\
Multi-Modal &\XSolidBrush  &\XSolidBrush&\XSolidBrush&\XSolidBrush&\Checkmark \\
\bottomrule
\end{tabular}}
\label{tab:datastat}
\end{table}

\noindent\textbf{Difficulty Measure of MESED} 

\begin{figure}[th]
\centering
\setlength{\abovecaptionskip}{0.5em}
\subfigure[Semantic class schema for MESED.] 
{ \label{fig:schama}
\includegraphics[width=0.486\columnwidth]{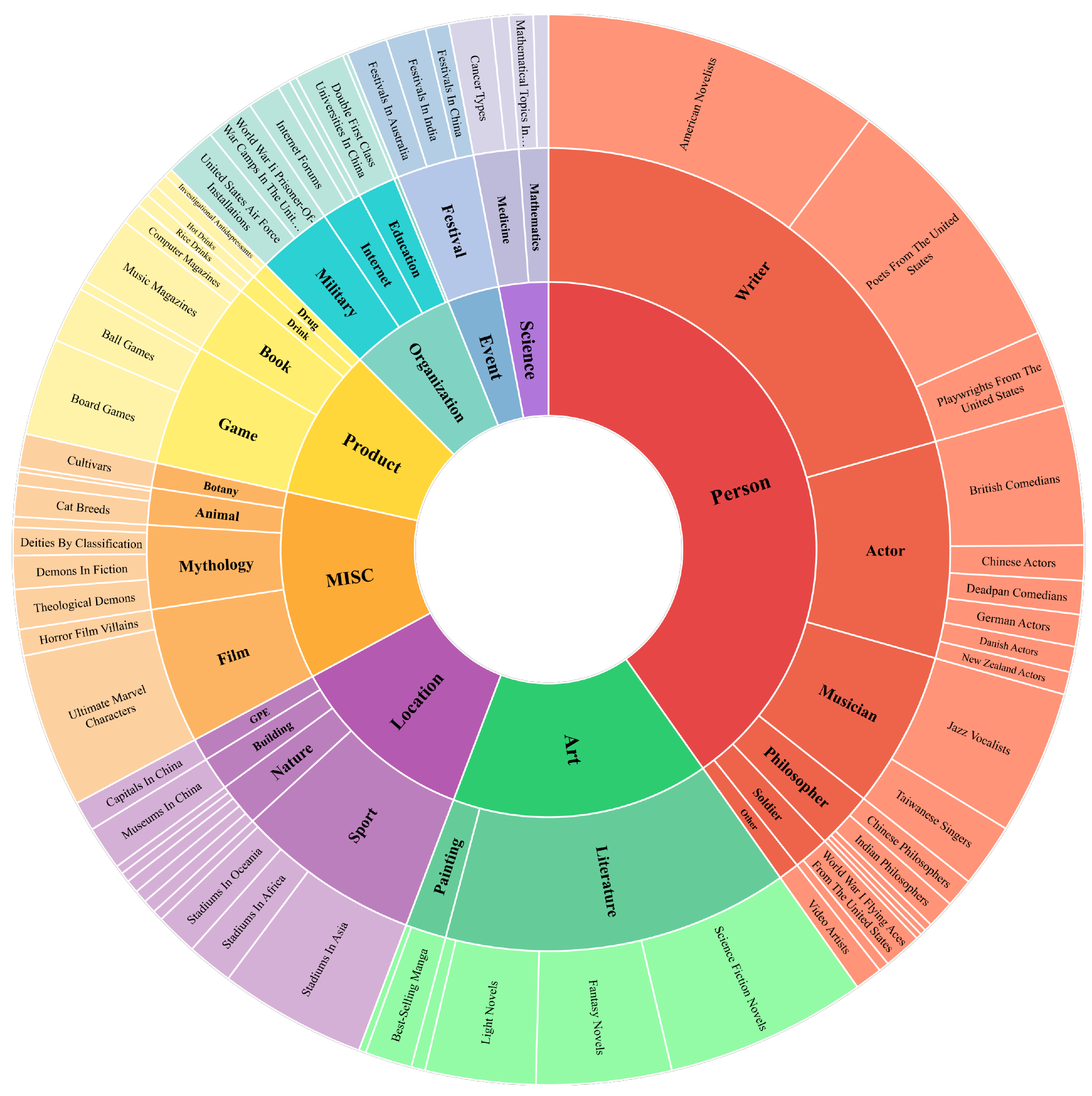} 
} 
\subfigure[Heat map visualization of semantic class similarity.] 
{ \label{fig:heatmap}
\includegraphics[width=0.46\columnwidth]{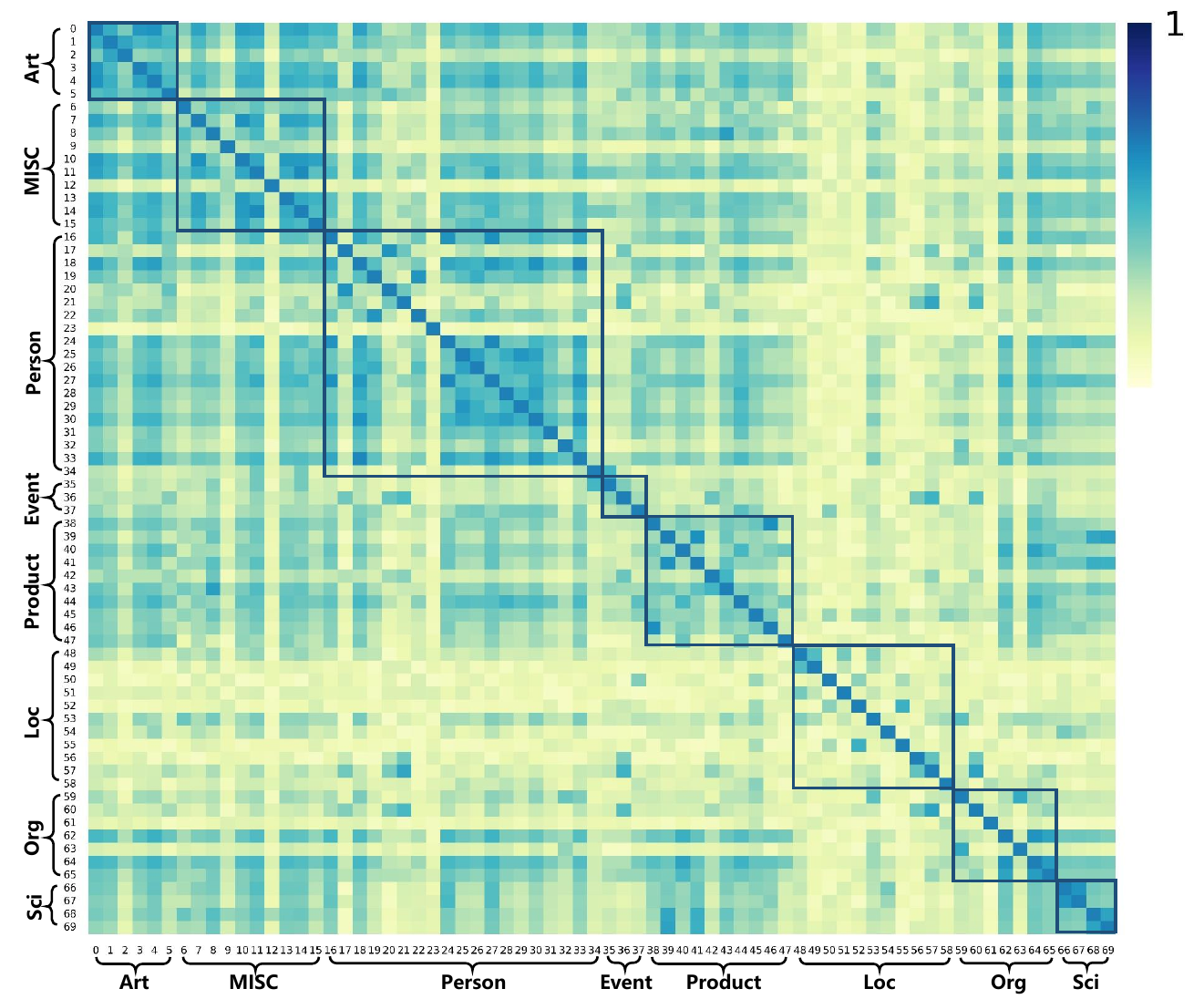} 
} 
\caption{Difficulty measure of MESED.} 
\label{tab:sensitive} 
\end{figure}

We ensured that the MESED was challenging from multiple perspectives:

Firstly, we meticulously designed the schema of semantic classes, as depicted in Figure \ref{fig:schama}, which consists of three layers. The first and second layers encompass 8 and 26 coarse-grained semantic classes, respectively, and the last layer contains 70 fine-grained semantic classes. Fine-grained semantic classes that belong to the same parent class have semantic overlap. Additionally, some semantic classes even have overlapping target entities, such as \textit{playwrights} and \textit{poets}, making them hard negative semantic classes for each other. This design improves the confusion of semantic class. We believe that multi-modal information can compensate for the semantic confusion of a single text modality.

Secondly, we included entities sharing words with the target entities obtained through the BM25-based Wikipedia search engine, as hard negative entities in the candidate word list. We argue that normal PLMs tend to inaccurately extend entities that have tokens in common with the seed entities while ignoring semantic mismatches due to the nature of masked language modeling.

Thirdly, we assessed the model's ability to expand synonymous entities by obtaining the entity's synonyms via Wikidata SPARQL\footnote{https://query.wikidata.org/} and replacing a portion of the entity with synonyms having an edit distance greater than 5 from it.

Furthermore, the heat map in Figure \ref{fig:heatmap} illustrates the cosine similarity between each fine-grained semantic class (represented by the average embedding of the entities). We observed that the similarity between fine-grained semantic classes contributing to the same coarse-grained semantic class (enclosed in dark squares) is higher, indicating greater semantic confusion among them.

\section{Methods}
\subsection{Overall Framework}
\begin{figure}[htb]
    \centering
    \includegraphics[width=0.98\columnwidth]{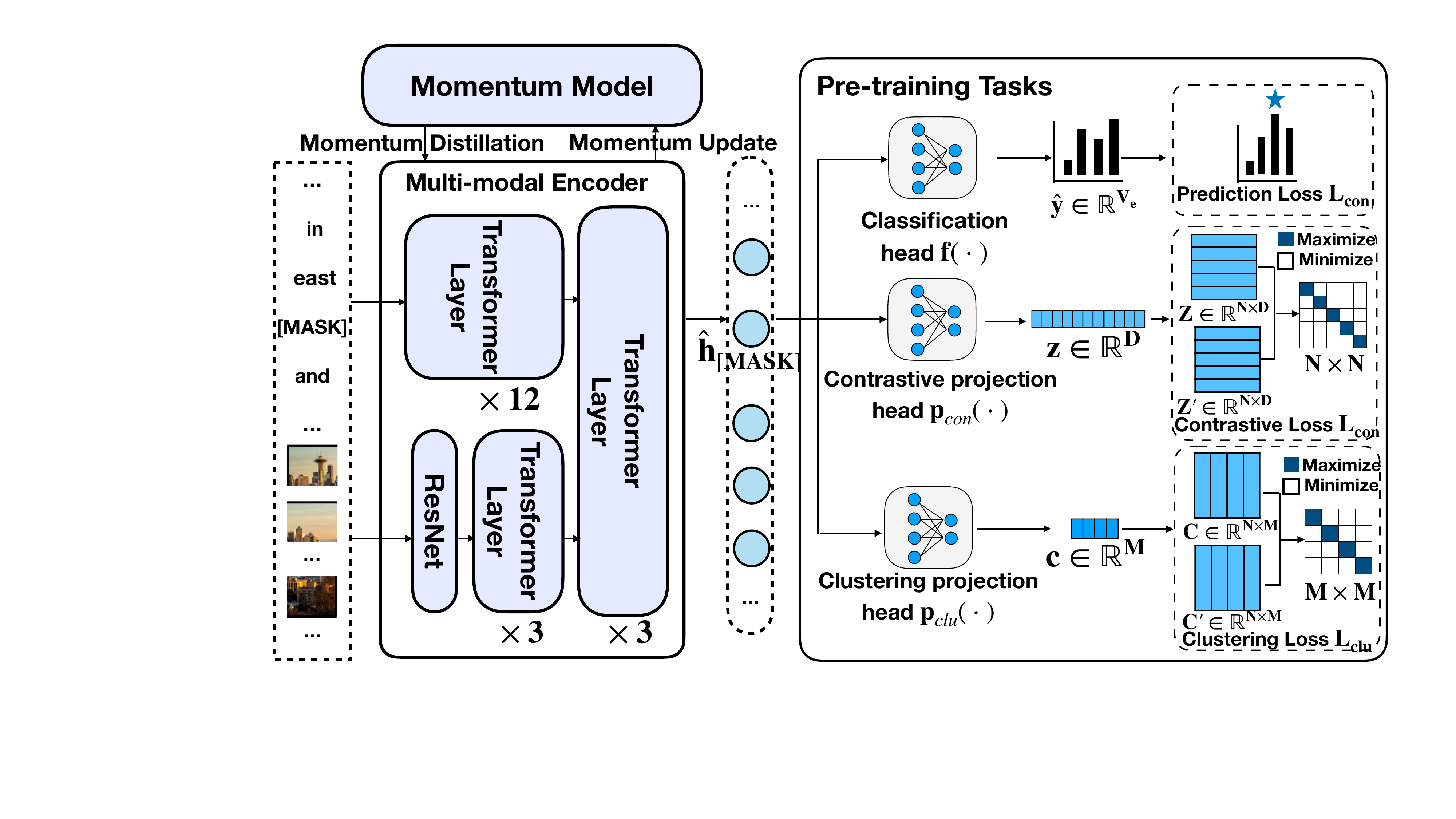}
    \caption{
    The pipeline of the multi-modal entity representation phase.
    }
    \label{fig:method}
\end{figure}
In this section, we describe the proposed MultiExpan method for MESE task, which expands the initial entity set with multi-modal contexts. Inspired by the previous ProbExpan \cite{li2022contrastive}, we divide MultiExpan into two steps: multi-modal entity representation phase and entity expansion phase. In the first phase, we design a multi-modal entity-level encoder whose output is the probability distribution of masked span over candidate entities. The entity is represented as the average of the predicted entity distributions for all sentences containing it. A multi-modal masked entity prediction task and three auxiliary self-learning pre-training tasks are proposed to refine the entity representation. In the second phase, MultiExpan obtains the target entities according to the similarities of the probabilistic representation of the entities. MultiExpan bears similarities to ProbExpan. We note that MultiExpan is proposed to provide a robust multi-modal baseline and to explore the effectiveness of different pre-training tasks.

\subsection{Multi-modal Entity Representation}
Multi-modal encoder first processes text and images separately with self-attention Transformer, then combines them for deep cross-modal interaction.

\textbf{Text}  First, we handle the text information. For the masked entity probability distribution pre-training model, we replace entity mentions in sentences with [MASK] to construct the inputs for text modality. Concerning the contextual text $T=\{w_1,w_2,...,w_{L_1}\}$ with masked entity mention, we directly use 12 layers of Transformer \cite{vaswani2017attention} initialized by $\text{BERT}_{\text{BASE}}$ \cite{kenton2019bert} to obtain the textual context's embeddings:
\begin{equation}
    \hat{W} = \{\hat{w}_1, \hat{w}_2, ..., \hat{w}_{L_1}\}=\text{BERT}_{\text{BASE}}(T)
\end{equation}
where $L_1$ is the max length of tokens in the sentences.

\textbf{Image}  Second, we deal with the image information. Different from the regional features and grid features widely used in the field of image feature extraction, the patch features we adopt are simple yet efficient. We transform each image into a fixed shape and determine the size of each patch, divide each image into 36 patches $I=\{i_1,i_2,..., i_{L_2}\}$, and use the backbone Resnet to extract patch features:
\begin{equation}
\{v_1, v_2, ..., v_{L_2}\}=Flat(Resnet(I))
\end{equation}
where $L_2$ is the number of patches and $Flat(\cdot)$ indicates the flatting function that reshapes the patch features extracted from Resnet into one dimensional.

Since the patch features will cause the loss of position information during segmentation, we add a learnable position embedding $P=\{p_1,p_2,...,p_{L_2}\}$ in order to mark the position information of each patch. Both patch features and position embeddings are combined through pair-wise add.

Finally, we build a 3-layer transformer architecture as image encoder in the visual information processing:
\begin{equation}
\hat{V} = \{\hat{v}_1, \hat{v}_2, ..., \hat{v}_{L_2}\}=Encoder_V(Flat(Resnet(I))\oplus P)
\end{equation}

\textbf{Cross-modal fusion}
After obtaining the information of the two modalities, the hidden states $\{h_1, h_2, ..., h_L\}$ are obtained through the concatenation of text features and visual features: $concat(\hat{W},\hat{V})$. Then we feed it into a 3-layer transformer for interaction and fusion between modalities so that the image-text pairs are fully aligned:
\begin{equation}
    \{\hat{h}_1, \hat{h}_2, ..., \hat{h}_L\}=Encoder_{cross}(\{h_1, h_2, ..., h_L\})
\end{equation}
where $L=L_1+L_2$ and the structure of the transformer is the same as the above-mentioned visual encoder.

A classification head $\mathbf{f}$ is attached behind the multi-modal encoder. After getting the hidden state of the mask position, the embedding vector is transformed into the probability distribution of the masked entity over the possible candidate entities by MLP and Softmax function:
\begin{equation}
    \hat{y} = \mathbf{f}(\hat{h}_{[MASK]})=  Softmax(MLP(\hat{h}_{[MASK]})), \hat{y}\in\mathbb{R}^{V_e}
\end{equation}
in which $V_e$ is the size of candidate entities vocabulary.

For the training of multi-modal encoder, four self-supervised pre-training objectives are adopted: masked entity prediction loss, contrastive learning loss, clustering learning loss, and momentum distillation loss. Eventually, the multi-modal encoder iteratively optimizes the four objectives.

\textbf{Masked entity prediction loss}
With respect to the masked entity prediction task, the model takes images and the masked sentences as input and obtains the entity probability distribution $\hat{y}$ of the masked position as described above. Cross-entropy loss with label smoothing is applied to allow the model to learn the underlying semantics of entities:
\begin{equation}\label{eq:mep}
\begin{split}
    \mathcal{L}_{mask} = -\frac{1}{N} \sum_i^{N} \sum_{j}^{V_e} y_{i}[j]\cdot(1-\eta)\cdot log(\hat{y}_{i}[j]) \\ +(1-y_{i}[j])\cdot \eta \cdot log(1-\hat{y}_{i}[j])
\end{split}
\end{equation}
where the ground truth $y$ is the one-hot vector and $N$ is the batch size. $\eta$ is the smoothing factor that prevents entities sharing semantics with the target entity from being overly suppressed.

\textbf{Contrastive learning loss}
Contrastive learning provides clearer semantic boundaries of semantic classes through drawing the representation of the same semantic class entities closer and the representation of different semantic class entities further apart~\cite{li-etal-2022-past, li-etal-2022-learning-dictionary}. We generate the positive and negative entities for each semantic class from the expanded list obtained in the previous iteration. The entities ranked in the top $K_{pos}$ positions are defined as positive entities, while the entities ranked from $L_{neg}$ to $U_{neg}$ are considered negative entities. The samples from positive/negative entities are paired to form positive/negative sample pairs. For a mini-batch of size $N$, each sample $x_{2i-1}$ forms $2N-1$ pairs with others, among which we pair ${x_{2i-1},x_{2i}}$ to be positive and define other $2N-2$ pairs to be negative.

Since directly performing contrastive learning on the hidden features $\hat{h}_{[MASK]}$ may result in information loss, we plugged in a two-layer MLP $\mathbf{p}_{con}(\cdot)$ behind multi-modal encoder to map the hidden features to a normalized subspace via $z_i=\mathbf{p}_{con}(\hat{h}_{[MASK]})$, where $z_i\in\mathbb{R}^{D}$ and $D$ is the dimension of subspace. The pair-wise similarity is measured by dot product:
\begin{equation}
    s(z_i,z_j)=z_i \cdot z_j^\top, i,j\in[1,2N]
\end{equation}
The contrastive learning loss that concentrates on hard negative entities is applied. For a given sample $z_i$ (suppose it forms a positive pair with $z_j$), the loss is defined as:
\begin{equation}\label{eq:con1}
    l_i =- \log \frac{e^{s(z_i,z_j) / t}}{e^{s(z_i,z_j) / t} + R_{i}^{-}},
\end{equation}

\begin{equation}\label{eq:con2}
    R_{i}^{-} = \max(\frac{-(2N-2)\cdot \tau\cdot e^{s(z_i,z_j) / t} +  \widetilde{R_{i}^{-}}}{1-\tau^{+}}, e^{-\frac{1}{t}})
\end{equation}
\begin{equation}\label{eq:con3}
    \widetilde{R_{i}^{-}} = \frac{(2N-2)\sum_{k: k \neq i \neq j} e^{(1+\beta)\cdot s(z_i,z_k) / t} }{\sum_{k: k \neq i \neq j} e^{\beta \cdot s(z_i,z_k) / t}}
\end{equation}
where $\tau, \beta, t$ are hyperparameters, representing class prior probability, hard negative entity concentration level, and temperature. The contrastive loss is computed across all samples in the batch:
\begin{equation}
    \mathcal{L}_{con} = \sum_{i=1}^{2N} l_i
\end{equation}

\textbf{Clustering learning loss}
Similar to contrastive learning, clustering learning attracts positive semantic class pairs and repels negative semantic class pairs.
We employ an alternative projection head, denoted as $\mathbf{p}_{clu}$, to map the input sample $x_i$ onto a semantic class subspace, resulting in $c_i = \mathbf{p}_{clu}(\hat{h}{[MASK]})$. The dimension $M$ of $c_i$ corresponds to the number of clusters, namely the number of target semantic classes. Each element of the feature indicates the probability that it belongs to a particular semantic class. We posit that a semantic class can be characterized by the probabilistic responses of a batch of entities towards it. Formally, let $C = [c_1, \cdots,c_{2i-1},\cdots, c_{2N-1}]\in\mathbb{R}^{N\times M}$ denotes the class probability distribution under samples $\{x_1,\cdots,x_{2i-1},\cdots,x_{2N-1}\}$, and  $C^{\prime} = [c_2, \cdots,c_{2i},\cdots, c_{2N}]$ for samples $\{x_2,\cdots,x_{2i},\cdots,x_{2N}\}$. 
The positive clustering pairs are formed by the semantic classes represented by the same columns of matrices $C$ and $C^{\prime}$, due to the fact that the entities $x_{2i-1}$ and $x_{2i}$, corresponding to each element of these column vectors, are positive sample pairs originating from the same semantic class. For brevity, we denote the $i$-th column of $C$ as $\hat{c}_{2i-1}$ and $\hat{c}_{2i}$ for the $i$-th column of $C^{\prime}$. Similarly, dot product is adopted to quantify the similarity between $\hat{c}_i$ and $\hat{c}_j$:
\begin{equation}
    \hat{s}(\hat{c}_i,\hat{c}_j)=\hat{c}_i^\top \cdot \hat{c}_j, i,j\in[1,2M]
\end{equation}
For each semantic class $\hat{c}_i$, the clustering loss $\hat{l}_i$ is computed in the same way as contrastive loss defined in Equation \eqref{eq:con1}-\eqref{eq:con3}, which distinguishes $\hat{c}_i$ from other $2M-2$ semantic classes except its positive counterpart $\hat{c}_j$. The clustering loss is finally calculated as:
\begin{equation}
    \mathcal{L}_{clu} = \sum_{i=1}^{2M} \hat{l}_i
\end{equation}

\textbf{Momentum distillation loss}
The image-sentence pairs in our MESED are collected from the web, often accompanied by noise, which causes the collected images may be weakly related to the sentences, or the extended entities belonging to the semantic class are not included in ground truth. To alleviate the above problems, we introduce momentum distillation learning. During training, a momentum version of the model is slowly updated by exponentially shifting the momentum factor $m$: $\theta_{\mathrm{t}} \leftarrow m \theta_{\mathrm{t}}+(1-m) \theta_{\mathrm{s}}$ and the momentum model is used to generate pseudo-labels as additional supervision, preventing the student model overfitting to noise.

The momentum distillation loss is expressed as the KL divergence between the pseudo entities probability distribution $\widetilde{y}$ generated by the momentum model and the predicted $\hat{y}$ of the multi-modal encoder at current iteration:
\begin{equation}
    \mathcal{L}_{mod} = -\sum_{i=1}^m \widetilde{y_i} log(\widetilde{y_i}) - \widetilde{y_i} log(\hat{y_i}))
\end{equation}

\subsection{Entity Expansion}
The entity is represented as the average of the predicted entity distributions for all sentences containing it. The semantic class is represented by the weighted average of entities in current expansion set and the weight is dynamically maintained by window search algorithm. In this way, candidate entities with similar distribution are placed in the current set measured by KL divergence. When the number of entities in current set reaches the target size, the entity re-ranking algorithm is performed to refine the final ranking list.

As the expansion process is not the focus of this work, we use window search and entity re-ranking algorithm from the ProbExpan and will not repeat them here. Please refer to \cite{li2022contrastive} for more details.



\section{Experiments}
\subsection{Experiment Setup}
\noindent\textbf{Compared Methods} We compare three categories of models, the first is the traditional text-based ESE approach, including \textbf{SetExpan}~\cite{shen2017setexpan}, \textbf{CaSE}~\cite{yu2019corpus}, \textbf{CGExpan}~\cite{zhang2020empower}, \textbf{ProbExpan}~\cite{li2022contrastive} and \textbf{GPT-3.5}. Of the above models, SetExpan, CaSE are the traditional statistical probability-based approaches, and CGExpan and ProbExpan are the most advanced methods based on pre-trained language model BERT. We also evaluated vision-based models: \textbf{VIT} \cite{dosovitskiy2020image}, \textbf{BEIT} \cite{bao2021beit} and image encoder of CLIP (\textbf{CLIP-IMG}). For multi-modal expansion, we explored multi-modal models with different structures comprising \textbf{CLIP} \cite{radford2021learning} and \textbf{ALBEF} \cite{li2021align}. Both the above-mentioned vision-based and multi-modal models are further pre-trained via entity prediction tasks, analogous to the method defined in Equation \eqref{eq:mep}. A detailed description of baselines is in Appendix.

\noindent\textbf{Evaluation Metrics} The objective of ESE is to expand ranked entity list ranked based on their similarity to given seed entities in descending order. Two widely used evaluation metrics, MAP@$K$ and P@$K$, are employed, as also utilized in previous research \cite{zhang2020empower, li2022contrastive, yan2020global}. The MAP@$K$ metric is computed as follows:
\begin{equation}
    \mbox{MAP@}K=
    \frac{1}{|Q|}\sum_{q \in Q}
    \mbox{AP}_K(R_q,G_q)
\end{equation}
Here, $Q$ is the collection for each query $q$. $\mbox{AP}_K(R_q,G_q)$ denotes the average precision at position $K$ with the ranked list $R_q$ and ground-truth list $G_q$. P@$K$ is the precision of the top-$K$ entities. In the experiment, queries with |Seed|=3 and 5 are evaluated separately.

\subsection{Main Experiment}

\begin{table}[]
\small
\centering
\renewcommand\arraystretch{1}
\setlength\tabcolsep{4.3pt}
\caption{Main experiment results. Text-based, vision-based, and multi-modal expansion methods are evaluated.}
\scalebox{0.75}{
\begin{tabular}{clccccccccc}
\toprule
\multirow{3}{*}{\textbf{Modality}} & \multirow{3}{*}{\textbf{Method}} & \multicolumn{9}{c}{\textbf{|Seed|=3}}                                                                                                                   \\ \cmidrule{3-11} 
                                   &                                  & \multicolumn{4}{c}{\textbf{MAP}}                           & \multicolumn{4}{c}{\textbf{P}}                             & \multirow{2}{*}{\textbf{Avg}} \\ \cmidrule{3-10}
                                   &                                  & \textbf{@10} & \textbf{@20} & \textbf{@50} & \textbf{@100} & \textbf{@10} & \textbf{@20} & \textbf{@50} & \textbf{@100} &                               \\ \midrule
\multirow{6}{*}{T}        & SetExpan                & 26.10 & 20.98 & 15.83 & 13.91 & 34.25 & 29.58 & 24.25 & 22.96 & 23.48                \\ \cmidrule{2-11} 
                          & CaSE                    & 27.71 & 20.93 & 14.63 & 12.02 & 36.85 & 30.57 & 24.83 & 23.63 & 23.90                \\ \cmidrule{2-11} 
                          & CGExpan                 & 38.89 & 32.51 & 24.69 & 21.06 & 45.85 & 39.85 & 33.19 & 32.80 & 33.61                \\ \cmidrule{2-11} 
                          & GPT-3.5                 & 31.10 & 24.73 & 19.20 & 17.07 & 37.65 & 31.35 & 26.08 & 25.11 & 26.54                \\ \cmidrule{2-11} 
                          & GPT+Name                & 42.12 & 35.32 & 26.83 & 23.21 & 52.32 & 41.23 & 35.89 & 35.73 & 36.58                \\ \cmidrule{2-11} 
                          & ProbExpan               & 65.47 & 57.50 & 43.96 & 40.73 & 71.30 & 64.35 & 55.73 & 51.99 & 56.38                \\ \midrule
\multirow{3}{*}{V}        & VIT                     & 65.02 & 55.94 & 41.89 & 32.40 & 67.95 & 59.53 & 46.08 & 36.94 & 50.72                \\ \cmidrule{2-11} 
                          & BEIT                    & 68.45 & 58.58 & 43.59 & 33.69 & 71.70 & 62.13 & 47.60 & 37.66 & 52.93                \\ \cmidrule{2-11} 
                          & CLIP-IMG                & 66.39 & 57.04 & 41.72 & 32.42 & 68.85 & 60.90 & 45.79 & 36.81 & 51.24                \\ \midrule
\multirow{4}{*}{T+V}      & CLIP                    & 76.41 & 65.75 & 49.58 & 40.08 & 79.20 & 69.53 & 53.10 & 43.66 & 59.66                \\ \cmidrule{2-11} 
                          & ALBEF                   & 83.55 & 75.46 & 63.02 & 54.47 & 86.60 & 79.15 & 68.03 & 61.12 & 71.43                \\ \cmidrule{2-11} 
                          & Ours (MEP)        & 86.07 & 79.18 & 67.66 & 58.91 & 89.10 & 82.85 & 72.13 & 65.17 & 75.13                \\ \cmidrule{2-11} 
                          & Ours (Full)       & 91.44 & 86.85 & 76.86 & 63.34 & 93.60 & 89.63 & 80.37 & 67.15 & 81.16 \\ \midrule \midrule
\multirow{3}{*}{\textbf{Modality}} & \multirow{3}{*}{\textbf{Method}} & \multicolumn{9}{c}{\textbf{|Seed|=5}}                                                                                                                   \\ \cmidrule{3-11} 
                                   &                                  & \multicolumn{4}{c}{\textbf{MAP}}                           & \multicolumn{4}{c}{\textbf{P}}                             & \multirow{2}{*}{\textbf{Avg}} \\ \cmidrule{3-10}
                                   &                                  & \textbf{@10} & \textbf{@20} & \textbf{@50} & \textbf{@100} & \textbf{@10} & \textbf{@20} & \textbf{@50} & \textbf{@100} &                               \\ \midrule
\multirow{6}{*}{T}        & SetExpan                & 25.99 & 20.64 & 15.20 & 13.51 & 34.90 & 29.93 & 24.26 & 23.29 & 23.47                \\ \cmidrule{2-11} 
                          & CaSE                    & 32.01 & 24.63 & 17.99 & 14.58 & 41.50 & 34.75 & 28.83 & 27.03 & 27.67                \\ \cmidrule{2-11} 
                          & CGExpan                 & 38.86 & 31.49 & 23.54 & 20.23 & 45.55 & 38.28 & 31.88 & 32.15 & 32.75                \\ \cmidrule{2-11} 
                          & GPT-3.5                 & 31.79                & 25.46                & 20.12                & 19.94                & 39.40                & 33.13                & 28.67                & 30.45                & 28.62              \\ \cmidrule{2-11} 
                          & GPT+Name               & 42.32                & 36.48                & 25.76                & 22.36                & 52.94                & 42.10                 & 34.68                & 35.12                & 36.47               \\ \cmidrule{2-11} 
                          & ProbExpan               & 66.29 & 59.31 & 48.90 & 42.51 & 73.15 & 66.78 & 58.51 & 54.54 & 58.75                \\ \midrule
\multirow{3}{*}{V}        & VIT                     & 62.29 & 55.43 & 41.30 & 31.54 & 68.20 & 58.93 & 45.61 & 35.91 & 49.90                \\ \cmidrule{2-11} 
                          & BEIT                    & 70.14 & 59.04 & 43.08 & 33.21 & 73.45 & 62.93 & 47.25 & 37.17 & 53.28                \\ \cmidrule{2-11} 
                          & CLIP-IMG                & 67.67 & 57.28 & 41.41 & 31.86 & 70.40 & 60.80 & 45.25 & 35.94 & 51.33                \\ \midrule
\multirow{4}{*}{T+V}      & CLIP                    & 77.37 & 65.92 & 49.01 & 39.05 & 79.80 & 69.48 & 52.41 & 42.50 & 59.44                \\ \cmidrule{2-11} 
                          & ALBEF                   & 85.04 & 76.25 & 62.45 & 53.64 & 87.80 & 79.70 & 67.37 & 60.06 & 71.54                \\ \cmidrule{2-11} 
                          & Ours (MEP)        & 87.77 & 79.96 & 67.24 & 57.62 & 90.90 & 83.55 & 71.41 & 63.41 & 75.23                \\ \cmidrule{2-11} 
                          & Ours (Full)       & 92.67 & 87.27 & 75.70 & 61.36 & 94.30 & 89.68 & 78.56 & 64.46 & 80.50      \\ \bottomrule
\end{tabular}}
\label{tab:main}
\end{table}

The results of the main experiment are presented in Table \ref{tab:main}, from which we observe that: (1) The multi-modal methods outperform the mono-modal methods in general. Remarkably, our MultiExpan (MEP) achieves superior performance solely by employing masked entity prediction task. The complete MultiExpan method achieves the best overall performance. Moreover, the full version of MultiExpan achieves optimal performance.

(2) In terms of the structure of multi-modal models, ALBEF and our MultiExpan exhibit deep modality interaction through the Transformer, which is better suited for the ESE task compared to the CLIP 's shallow modal interaction via dot product similarity calculation. These results indicate that deep modal interaction and fusion is a direction that can be explored in the future.

(3) In terms of the vision-based models, BEIT excels in leveraging finer-grained image semantics, such as object and background information, by pre-training on masked image modeling. In contrast to the VIT model which learns the overall image semantics through image classifying images in the Image Net dataset, BEIT demonstrates better results in entity understanding. Meanwhile, the image encoder of CLIP also captures richer semantics than the VIT model owing to its linkage with the text modality. However, relying solely on image modality does not suffice to produce satisfactory results, and the text modality still remains dominant.

(4) The increase of |Seed| does not necessarily translate to an enhancement in overall performance. More seeds can describe the semantic classes more precisely and retrieve some "must be correct" entities more safely, so MAP/P improves when K is small (=10,20). However, more seed entities mean a larger search space for semantic classes, necessitating a more meticulous analysis of common entity properties than the current model allows. This issue represents the persistent challenge of semantic drift that confronts ESE models, so MAP/P decreases when K is larger. Of course, increasing |Seed| helps disambiguate the query with entities belonging to multiple classes. Such as in the case of the semantic class \textit{Light Novel}, where some seed entities also are \textit{Manga}, increasing |Seed| makes a gain of 17.5\% average on all metrics.

(5) GPT-3.5 did not achieve satisfactory results, and was even inferior to unsupervised CGExpan. Through meticulous examination of GPT-3.5's performance on specific semantic classes, we discovered that the model struggled with complex classes (e.g., \textit{108 Martyrs of World War II}). We explicitly instructed GPT-3.5 to reason about the class names first, and then expand based on them. This modification, named GPT+Name, exhibited a substantial improvement compared to GPT-3.5. This approach aligns with the idea of emerging chain-of-thought reasoning \citep{wei2022chain} for large language models, i.e., thinking step by step. We suggest future research to explore the combination of chain-of-thought and ESE tasks.

\subsection{Pre-training Tasks Analysis}

\begin{table}[]
\centering
\renewcommand\arraystretch{1}
\setlength\tabcolsep{3.8pt}
\caption{Comparison of different pre-training tasks.}
\scalebox{0.75}{
\begin{tabular}{lccccccccc}
\toprule
\multirow{2}{*}{\textbf{Model}} & \multicolumn{4}{c}{\textbf{MAP}}                           & \multicolumn{4}{c}{\textbf{P}}                             & \multirow{2}{*}{\textbf{Avg}} \\ \cmidrule{2-9}
                                 & \textbf{@10} & \textbf{@20} & \textbf{@50} & \textbf{@100} & \textbf{@10} & \textbf{@20} & \textbf{@50} & \textbf{@100} &                               \\ \midrule
MultiExpan (MEP)                               & 86.07        & 79.18        & 67.66        & 58.91         & 89.10        & 82.85        & 72.13        & 65.17         & 75.13                         \\ \midrule
+ Contrastive                               & 90.71        & 86.58        & 75.58        & 62.69         & 93.35        & 89.60        & 79.23        & 67.10         & 80.61                         \\ \midrule
+ Clustering                               & 89.10        & 82.83        & 70.85        & 60.48         & 91.65        & 86.05        & 74.75        & 65.92         & 77.70                         \\ \midrule
+ Distillation                               & 86.97        & 80.48        & 68.30        & 59.43         & 89.85        & 83.65        & 72.34        & 65.23         & 75.78                        \\ \midrule
MultiExpan (Full)                               & 91.44        & 86.85        & 76.86        & 63.34         & 93.60        & 89.63        & 80.37        & 67.15         & 81.16                        \\ \bottomrule
\end{tabular}}
\label{tab:ablation1}
\end{table}

We compared the effects of different pre-training tasks on MultiExpan. The masked entity prediction task enables the model to learn the underlying semantics of entities, which is further enhanced by the addition of three pre-training tasks. The results presented in Table \ref{tab:ablation1} demonstrate that each pre-training task confers a gainful effect on the model. Notably, we found that contrastive learning with hard negative entities yields the greatest performance improvement for the model, by providing clearer semantic boundaries. While clustering learning brings comparable gains to contrastive learning at MAP/P@K=10 and 20, it is less effective at larger K. This is because contrastive learning operates directly on entities and more directly aggregates target entities into tight clusters. In contrast, momentum distillation learning brings a smaller performance gain, which we believe is mainly attributed to its ability to prevent overfitting in the presence of noisy data. This observation underscores the high quality of the data provided by MESED, particularly the accurate annotation of entities in sentences.

Extensive experiments on the hyperparameters sensitivity of the pre-training tasks are presented in Appendix, demonstrating the robustness of MultiExpan to the parameters.

\subsection{Modality Analysis}
We also carry out analysis experiments on each modality to answer the following questions.

\noindent\textbf{Are the multiple modalities complementary?}

\begin{wrapfigure}{r}{0.26\textwidth}
        \includegraphics[width=0.2\textheight]{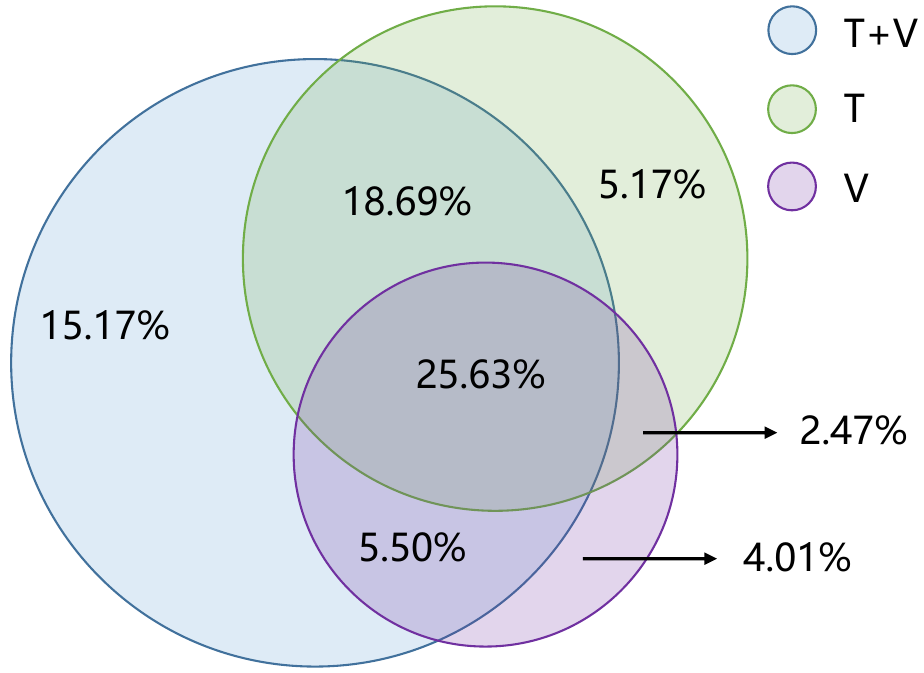}
        \caption{Venn diagram showing contribution of each modality.}
        \label{fig:venn}
\end{wrapfigure}
We present a Venn diagram illustrating the impact of different modalities on MESE, as depicted in Figure \ref{fig:venn}. T, V and T+V represent ProbExpan, BEIT and our MultiExpan respectively. The size of each circle corresponds to the proportion of the top 100 ranked entities that belong to the ground truth, and the intersection of the circles represents the overlap of entities. Our analysis shows that the textual modality still prevails over the visual modality. Whereas the visual modality is introduced as supplementary information, 15.17\% of the target entities in MultiExpan are sorted to a higher position, while 5.17\% of the entities that were originally correctly expanded are excluded, due to the image noise.

Additional case studies and detailed model performance evaluations for each semantic class can be found in Appendix.

\begin{table}[]
\centering
\renewcommand\arraystretch{1}
\setlength\tabcolsep{3.8pt}
\caption{Ablation study on modality absence.}
\vspace{-0.1cm}
\scalebox{0.75}{
\begin{tabular}{lccccccccc}
\toprule
\multirow{2}{*}{\textbf{Model}} & \multicolumn{4}{c}{\textbf{MAP}}                           & \multicolumn{4}{c}{\textbf{P}}                             & \multirow{2}{*}{\textbf{Avg}} \\ \cmidrule{2-9}
                                & \textbf{@10} & \textbf{@20} & \textbf{@50} & \textbf{@100} & \textbf{@10} & \textbf{@20} & \textbf{@50} & \textbf{@100} &                               \\ \midrule
MultiExpan (MEP)                       & 86.07        & 79.18        & 67.66        & 58.91         & 89.10        & 82.85        & 72.13        & 65.17         & 75.13                         \\ \midrule
pre-train w/o T                     & 65.97        & 57.87        & 42.84        & 33.39         & 70.45        & 62.50        & 48.85        & 39.70         & 52.70                         \\ \cmidrule{2-10} 
pre-train w/o V                     & 66.87        & 60.18        & 52.26        & 47.57         & 73.90        & 68.13        & 62.45        & 60.88         & 61.53       \\ \midrule
w/o $T_s$ and $T_c$               & 20.67        & 18.32        & 13.13        & 9.66          & 27.80        & 26.10        & 21.54        & 18.17         & 19.42                         \\ \cmidrule{2-10} 
w/o $T_s$                        & 20.75        & 18.43        & 13.57        & 9.92          & 27.50        & 25.88        & 21.54        & 18.34         & 19.49                         \\ \cmidrule{2-10} 
w/o $T_c$                        & 85.45        & 77.99        & 66.53        & 56.58         & 88.10        & 81.95        & 71.17        & 62.68         & 73.81                         \\ \cmidrule{2-10} 
w/o $V_s$ and $V_c$               & 58.99        & 50.36        & 40.38        & 35.60         & 64.05        & 56.53        & 48.37        & 47.09         & 50.17                         \\ \cmidrule{2-10} 
w/o $V_s$                        & 60.44        & 51.95        & 41.92        & 37.18         & 65.05        & 57.55        & 49.25        & 47.67         & 51.38                         \\ \cmidrule{2-10} 
w/o $V_c$                        & 84.79        & 76.94        & 64.55        & 55.92         & 87.90        & 81.18        & 69.76        & 63.07         & 73.01                         \\           \bottomrule       
\end{tabular}}
\label{tab:ablation-modal}
\vspace{-0.25cm}
\end{table}

\noindent\textbf{Is it better to have multi-modal contexts of both seed and candidate entities?} During the inference phase, we separately removed the textual and visual information from the candidate or seed entities in MultiExpan. The resulting performances are shown in the last 6 rows of Table \ref{tab:ablation-modal}, with subscripts indicating the operations performed on seeds (s) or candidates (c). Our results indicate that removing any part of the modal information for any part of the entities is detrimental to the overall performance. However, when a particular modal information was removed from seed entities, it caused a severe performance degradation, whereas removing modal information from candidate entities caused only a slight performance loss. These findings suggest that modeling the semantics of seed entity set is more crucial than modeling individual entities. Additionally, MultiExpan demonstrated a decrease in performance when we removed the input text or images separately during the pre-training phase, further demonstrating its ability to effectively utilize multimodal information.

\begin{table}[]
\caption{Model performance under different visual clues.}
\vspace{-0.1cm}
\scalebox{0.8}{
\begin{tabular}{lccc}
\toprule
\multirow{2}{*}{Visual Clues} & \multirow{2}{*}{Proportion} & \multicolumn{2}{c}{P@100} \\ \cmidrule{3-4} 
                              &                             & ProbExpan           & MultiExpan         \\ \midrule
Object                        & 46.3\%                      & 57.44       & 70.21       \\
Scene                         & 21.2\%                      & 67.44       & 72.09       \\
Property                      & 22.2\%                      & 66.66       & 80.00       \\
Others                        & 3.4\%                       & 61.90       & 76.19      \\ \bottomrule
\end{tabular}}
\label{tab:ablation:clue}
\vspace{-0.5cm}
\end{table}

\noindent\textbf{What visual clues are provided by the visual modality?} We randomly sample 200 entities and determine that images can provide essential visual clues, including (1) Objects, which can augment the limited textual information by depicting the entities themselves, (2) Scenes, which showcase the environment where the entity exists to differentiate between the target semantic class and the hard negative semantic class, e.g., indoor vs. outdoor, water vs. land, (3) Properties, which demonstrate the common traits of entities to align entities of the same class, such as appearance of \textit{Cats}, and (4) Other: Other important visual clues. We annotate 200 entity images with their corresponding visual clue types and assess MultiExpan's capacity to leverage different visual clues. As Table \ref{tab:ablation:clue} shows, all types of visual cues are beneficial to MESE, and visual modalities mainly supplement the textual information by highlighting objects in the images. In contrast, MultiExpan utilizes scenes to a lesser extent as they represent more abstract concepts. Examples of visual clues can also be found in Appendix.

\subsection{Ablation Experiment over Dataset}
\noindent\textbf{Dataset Difficulty} 
During the construction of MESED, we incorporated a variety of challenging entities. Our analysis, as shown in Table \ref{tab:ablation_diffi}, demonstrates that the performance of the model improves with the removal of each part of the entities. Notably, the hard negative semantic classes, which we specifically constructed for fine-grained semantic distinctions, posed significant difficulties for both MultiExpan and ProbExpan. On the other hand, removing synonym entities did not result in much performance change, possibly because some of the target synonym entities were not originally expanded by MultiExpan, making their removal insignificant. More importantly, we observed that these hard entities caused a much larger performance degradation for mono-modal ProbExpan, highlighting the importance of multi-modal information for all types of hard entities and strongly supporting our motivation.

\begin{table}[]
\centering
\renewcommand\arraystretch{1}
\setlength\tabcolsep{3.8pt}
\caption{Ablation experiments on dataset difficulty. Above: MultiExpan and Bottom: ProbExpan.}
\vspace{-0.1cm}
\scalebox{0.75}{
\begin{tabular}{lccccccccc}
\toprule
\multirow{2}{*}{\textbf{Dataset}} & \multicolumn{4}{c}{\textbf{MAP}}                           & \multicolumn{4}{c}{\textbf{P}}                             & \multirow{2}{*}{$\mathbf{\overline{\Delta}}$} \\ \cmidrule{2-9}
                                  & \textbf{@10} & \textbf{@20} & \textbf{@50} & \textbf{@100} & \textbf{@10} & \textbf{@20} & \textbf{@50} & \textbf{@100} &                               \\ \midrule
Full                               & 86.07        & 79.18        & 67.66        & 58.91         & 89.10        & 82.85        & 72.13        & 65.17         & -                             \\ \cmidrule{2-10} 
w/o polysemy                  & 86.84        & 79.68        & 67.88        & 59.53         & 89.25        & 83.05        & 72.86        & 65.43         & 0.43                          \\ \cmidrule{2-10} 
w/o hard neg                & 87.45        & 81.36        & 70.68        & 61.10         & 90.20        & 84.98        & 74.75        & 67.16         & 2.08                          \\ \cmidrule{2-10} 
w/o random neg                      & 88.91        & 81.98        & 69.16        & 61.85         & 91.05        & 84.25        & 73.06        & 65.97         & 1.90                          \\ \cmidrule{2-10} 
w/o synonym                             & 86.27        & 79.26        & 67.69        & 58.94         & 89.19        & 83.09        & 72.21        & 65.39         & 0.12                          \\ \midrule
Full                               & 65.47        & 57.50        & 43.96        & 40.73         & 71.30        & 64.35        & 55.73        & 51.99         & -                             \\ \cmidrule{2-10} 
w/o polysemy                  & 66.91        & 59.24        & 49.70        & 43.40         & 72.75        & 65.60        & 57.82        & 54.19         & 2.32                          \\ \cmidrule{2-10} 
w/o hard neg                & 71.82        & 64.35        & 52.64        & 47.70         & 76.10        & 70.18        & 63.66        & 57.56         & 6.62                          \\ \cmidrule{2-10} 
w/o random neg                      & 69.79        & 64.24        & 51.51        & 45.54         & 74.40        & 68.82        & 56.41        & 51.06         & 3.84                         \\ \cmidrule{2-10} 
w/o synonym                             & 65.90         & 57.69        & 44.33        & 40.86         & 71.63        & 64.77        & 56.13        & 52.20          & 0.31        \\ \bottomrule                 
\end{tabular}}
\label{tab:ablation_diffi}
\vspace{-0.25cm}
\end{table}

\noindent\textbf{Dataset Quality} We conduct ablation experiments on each stage of dataset construction to evaluate their impact on data quality. As Table \ref{tab:ablation22} illustrates, both human annotation and our proposed image re-ranking algorithm enhanced the quality of the image-sentence pair data, resulting in less interference from noise for the model. 
Furthermore, the two components of the image re-ranking algorithm, which assess image-sentence relevance (i\&s) and image-entity relevance (i\&e) respectively, are both essential to guarantee the informative value of the images while ensuring their abundance.

\begin{table}[]
\centering
\renewcommand\arraystretch{1}
\setlength\tabcolsep{3.8pt}
\caption{Ablation experiments on dataset quality.}
\vspace{-0.1cm}
\scalebox{0.75}{
\begin{tabular}{lccccccccc}
\toprule
\multirow{2}{*}{\textbf{Dataset}} & \multicolumn{4}{c}{\textbf{MAP}}                           & \multicolumn{4}{c}{\textbf{P}}                             & \multirow{2}{*}{\textbf{Avg}} \\ \cmidrule{2-9}
                                  & \textbf{@10} & \textbf{@20} & \textbf{@50} & \textbf{@100} & \textbf{@10} & \textbf{@20} & \textbf{@50} & \textbf{@100} &                               \\ \midrule
Full                & 86.07 & 79.18 & 67.66 & 58.91 & 89.10 & 82.85 & 72.13 & 65.17 & 75.13                \\ \midrule
w/o human                 & 84.38 & 77.19 & 65.90 & 57.28 & 87.55 & 81.20 & 70.66 & 64.47 & 73.58                \\ \midrule
w/o re-rank                 & 81.53 & 73.25 & 62.71 & 53.99 & 85.20 & 77.73 & 68.32 & 62.65 & 70.67      \\ \midrule
w/o re-rank i\&s               & 82.38 & 74.07 & 69.07 & 54.55 & 85.80 & 78.13 & 68.22 & 62.71 & 71.87                \\ \midrule
w/o re-rank i\&e               & 83.03 & 74.52 & 63.84 & 56.90 & 86.30 & 78.73 & 69.48 & 64.84 & 72.21                \\ \bottomrule
\end{tabular}}
\label{tab:ablation22}
\vspace{-0.4cm}
\end{table}


\section{Conclusion}
In this paper, we introduce a novel task called Multi-modal Entity Set Expansion (MESE), which aims to leverage multiple modalities to represent and expand entities. The MESED dataset is constructed which is the first multi-modal dataset for ESE with fine-grained semantic classes and hard negative entities. In addition, A powerful multi-modal model MultiExpan is proposed which is pre-trained on four multimodal pre-training tasks. MultiExpan achieves state-of-the-art results compared to other mono/multi-modal models. In the future, we will investigate the applicability of generative PLMs, such as GPT-4, in addressing MESE task. MESED can also serve as a reliable benchmark for assessing the multi-modal entity understanding capacities of large PLMs.

\bibliographystyle{ACM-Reference-Format}
\bibliography{sample-base}

\newpage
\clearpage
\appendix
\label{sec:appendix}

\section{ESE-related Multi-modal Resources}
\label{Appendix_A}
When exploring the interpretation of MESE, numerous multi-modal entity-level tasks are associated with MESE, and related publicly available datasets are presented by previous studies. Multi-modal Named Entity Recognition (MNER) \cite{yu2020improving,zhang2021multi} seeks to enhance text-based NER by utilizing images as supplementary inputs. MNER benchmark datasets are typically constructed from social media platforms, such as Twitter \cite{zhang2018adaptive,lu2018visual} and Snapchat \cite{moon2018multimodal}. As previously discussed, MNER datasets are limited in terms of entity types, and their granularities are insufficient to truly reflect the capabilities of the ESE model. Multi-modal Entity Linking (MEL) \cite{gan2021multimodal,zhang2021attention,wang2022wikidiverse} aims to link ambiguous mentions within a multimodal context to unambiguous entities in a given knowledge base. Online multimedia sites, like news and movie sites, are common sources of MEL data. For example, the M3EL dataset \cite{gan2021multimodal} collects movie reviews and corresponding images from the movie review sites IMDB and TMDB. The movie characters in the reviews are manually labeled and aligned with standard entities in Wikipedia. In the MEL dataset, data is typically presented as image-sentence pairs where sentences are marked with entity boundaries. Despite containing multi-modal contextual information about entities, they are not organized by semantic class. As described in Section \ref{sec:data_collect}, the induction of fine-grained semantic classes over large-scale entities is quite challenging. Hence, all these multi-modal datasets were not used to create MESE datasets.

\section{Implementation details of MultiExpan}
\label{Appendix_B}
The MultiExpan model employs a multi-modal encoder to encode input entities and their corresponding multi-modal contexts using two mono-modal encoders and a cross-modal interaction module, and decodes them with MLP-based classification heads. The multi-modal encoder is composed of a stacked Transformer layer \cite{vaswani2017attention}, which comprises multi-head attention and feed-forward networks. For textual input, sentences are tokenized using the WordPiece tokenizer \cite{wu2016google} and then fed into a 12-layer Transformer initialized with $BERT_{BASE}$ \cite{kenton2019bert} weights. To optimize training efficiency and preserve semantic knowledge learned by BERT, the first 11 layers of the text encoder are frozen, and only the last layer is fine-tuned. For images, the ResNet processes images into L features of patches, which are further encoded using a 3-layer Transformer. We designed the text encoder with more layers than the image encoder because we consider text to be a more abstract symbol and require deeper processing. The cross-modal interaction module is also composed of three layers of Transformer.

During training, we start by combining masked entity prediction loss and momentum distillation loss to predict entity probability distributions over masked spans. Consequently, we obtain positive/negative samples from entity expansion results and use them to construct positive/negative pairs with different formats, considering the different forms of different pretraining objectives. We then use contrastive learning and clustering learning to train MultiExpan and iterate the above process.

To select hyperparameters for the experiment, we set the learning rate and weight decay of MultiExpan to $10^{-5}$ and 0.01, respectively. Additionally, the \{smoothing factor $\eta$, lower bound of negative $L_{neg}$ entities, upper bound of negative entities $U_{neg}$, clustering number $M$, momentum factor $m$\} are set to \{0.075, 170, 200, 41, 0.99\}. The output dimensions of the two auxiliary projecting heads for contrastive learning and clustering learning are 128 and 41, respectively.

\section{Details of baselines}
\label{Appendix_c}
We compare three categories of models. The first is the traditional text-based ESE approach:

(1) \textbf{SetExpan}~\cite{shen2017setexpan}: An iterative expansion framework with a context feature selection method based on ensemble ranking.

(2) \textbf{CaSE}~\cite{yu2019corpus}: An unsupervised corpus-based set expansion framework that leverages lexical features as well as distributed representations of entities.

(3) \textbf{CGExpan}~\cite{zhang2020empower}: A framework that uses probing Hearst template \cite{hearst1992automatic} and BERT \cite{kenton2019bert} to generate candidate class names, and guide the entity selection process with the selected class names.
 
(4) \textbf{ProbExpan}~\cite{li2022contrastive}: Current state-of-the-art method which proposes an entity-level masked language model and empowered it to better handle hard negative entities with contrastive learning.

(5) \textbf{GPT-3.5}: GPT-3.5 is a mono-modal version of GPT-4. As OpenAI only provides a web demo with restricted GPT-4, we chose to evaluate GPT-3.5, which also achieved impressive zero-sample comprehension on the natural language understanding task. As a generative model, we constructed prompt templates with seed entities, allowing the model to generate more entities of the same semantic class on request. 
Furthermore, prior to expanding the seeds based on their respective classes, we utilized GPT-3.5 to generate initial class names, employing the method denoted as GPT+Name.

Of the above models, SetExpan, CaSE are the traditional statistical probability based approaches, and CGExpan and ProbExpan are the most advanced methods based on pre-trained language model BERT. We also evaluated vision-based models:

(1) \textbf{VIT} \cite{dosovitskiy2020image} is a Transformer-based pre-trained vision model that is pre-trained on large-scale image recognition data to learn image-level semantics.

(2) \textbf{BEIT} \cite{bao2021beit} first tokenizes the input image and then randomly masks patches of the image, allowing the model to recover the original image based on the mask. We believe BEIT can learn the semantics of the image at a finer granularity.

(3) \textbf{CLIP-IMG} utilizes image encoder of CLIP described below.

For multi-modal expansion, we explored multi-modal models with different structures:

(1) \textbf{CLIP} \cite{radford2021learning} learns individual encoders for each modality and only allows shallow interaction between modalities via dot product. During pre-training, CLIP performs contrastive learning on large-scale image-sentence pair datasets.

(2) \textbf{ALBEF} \cite{li2021align} The primary structural difference between ALBEF and CLIP resides in the capability of ALBEF to enable cross-modal interactions through Transformer structure at a deeper level. ALBEF's pre-training encompasses multiple tasks, such as masked language modeling and image-text matching, among others.

Both the above-mentioned vision-based and multi-modal models are further pre-trained via entity prediction tasks, analogous to the method defined in Equation \eqref{eq:mep}.

\section{More Analysis Experiments}
\label{Appendix_D}
\subsection{Performance on different semantic classes}
Table \ref{tab:each_class_map} displays a detailed evaluation of the mono-modal ProbExpan and our multi-modal MultiExpan across various fine-grained semantic classes. Our results demonstrate that MultiExpan surpasses ProbExpan with an average improvement of 15.34\% on nearly all semantic classes, particularly those with distinct visual characteristics, such as \texttt{Actors} and \texttt{Animals}, where MultiExpan's performance gain is particularly evident, aligning with our intuitive expectations. However, MultiExpan exhibits a decline in performance for \texttt{Mathematics} semantic classes. Case study in Appendix \ref{append_sec_case} reveals that the images for these classes primarily feature mathematical diagrams that pose a challenge for effective image exploitation, indicating the requirement for further exploration of strategies for leveraging image data.

\begin{table*}[]
\centering
\renewcommand{\arraystretch}{1.44}
\setlength\tabcolsep{3.8pt}
\caption{Performance on different semantic classes.}
\vspace{-0.2cm}
\scalebox{0.7}{
\begin{tabular}{llccccccc}
\toprule
\multicolumn{1}{l}{\multirow{2}{*}{\textbf{Coarse-grained Semantic Class}}} & \multicolumn{1}{l}{\multirow{2}{*}{\textbf{Fine-grained Semantic Class}}} & \multicolumn{3}{c}{\textbf{MAP@100}}                            & \multicolumn{3}{c}{\textbf{P@100}}                              & \multirow{2}{*}{\textbf{$\mathbf{\overline{\Delta}}$}} \\ \cmidrule{3-8}
\multicolumn{1}{c}{}                                                        & \multicolumn{1}{c}{}                                                      & \textbf{ProbExpan} & \textbf{MultiExpan} & \textbf{$\mathbf{\Delta}$} & \textbf{ProbExpan} & \textbf{MultiExpan} & \textbf{$\mathbf{\Delta}$} &                               \\ \midrule
\multirow{2}{*}{Literature}                                                 & 19th-century British Children Literature Titles                           & 3.34               & 23.46               & 20.12                & 7.33               & 31.33               & 24.00                 & 22.06                         \\ \cmidrule{2-9}
                                                                            & Light Novels                                                              & 7.38               & 43.24               & 35.86                & 22.80               & 55.40                & 32.60               & 34.23                         \\ \midrule
Painting                                                                    & Best-selling Manga                                                        & 17.42              & 35.44               & 18.02                & 24.88              & 43.72               & 18.84                & 18.43                         \\ \midrule
\multirow{2}{*}{Animal}                                                     & Cat Breeds                                                                & 58.22              & 80.25               & 22.03                & 61.05              & 80.26               & 19.21                & 20.62                         \\ \cmidrule{2-9}
                                                                            & National Animals                                                          & 6.11               & 35.77               & 29.66                & 28.65              & 43.78               & 15.13                & 22.40                         \\ \midrule
Botany                                                                      & Cultivars                                                                 & 22.07              & 46.06               & 23.99                & 52.00               & 55.33               & 3.33                 & 13.66                         \\ \midrule
\multirow{2}{*}{Mythology}                                                  & Demons in Fiction                                                         & 4.11               & 13.33               & 9.22                 & 17.14              & 20.57               & 3.43                 & 6.33                          \\ \cmidrule{2-9}
                                                                            & Theological Demons                                                        & 8.32               & 64.82               & 56.50                 & 16.14              & 67.95               & 51.81                & 54.16                         \\ \midrule
\multirow{2}{*}{Film}                                                       & Horror Film Villains                                                      & 39.37              & 46.50                & 7.13                 & 50.80               & 59.20                & 8.40                  & 7.77                          \\ \cmidrule{2-9}
                                                                            & Ultimate Marvel Characters                                                & 47.90               & 56.15               & 8.25                 & 59.80               & 57.00                & -2.80                 & 2.73                          \\ \midrule
Writer                                                                      & Playwrights From the United States                                        & 82.41              & 91.84               & 9.43                 & 85.80               & 92.00                & 6.20                  & 7.82                          \\ \midrule
Philosopher                                                                 & Chinese Philosophers                                                      & 80.54              & 86.86               & 6.32                 & 83.43              & 87.14               & 3.71                 & 5.02                          \\ \midrule
\multirow{2}{*}{Actor}                                                      & Chinese Actors                                                            & 28.95              & 71.33               & 42.38                & 47.44              & 81.16               & 33.72                & 38.05                         \\ \cmidrule{2-9}
                                                                            & Deadpan Comedians                                                         & 15.16              & 42.33               & 27.17                & 26.44              & 50.22               & 23.78                & 25.48                         \\ \midrule
Musician                                                                    & Jazz Vocalists                                                            & 97.68              & 99.20                & 1.52                 & 98.20               & 99.40                & 1.20                  & 1.36                          \\ \midrule
\multirow{2}{*}{Soldier}                                                    & 108 Martyrs of World War II                                               & 44.91              & 47.87               & 2.96                 & 47.41              & 51.85               & 4.44                 & 3.70                           \\ \cmidrule{2-9}
                                                                            & World War I Flying Aces From the United States                            & 31.49              & 48.72               & 17.23                & 33.17              & 49.02               & 15.85                & 16.54                         \\ \midrule
Other                                                                       & Video Artists                                                             & 57.67              & 85.55               & 27.88                & 58.46              & 86.77               & 28.31                & 28.10                         \\ \midrule
Festival                                                                    & Festivals in Australia                                                    & 60.05              & 70.38               & 10.33                & 67.16              & 71.64               & 4.48                 & 7.41                          \\ \midrule
\multirow{2}{*}{Game}                                                       & Ball Games                                                                & 75.26              & 98.71               & 23.45                & 78.80               & 98.80                & 20.00                 & 21.73                         \\ \cmidrule{2-9}
                                                                            & Board Games                                                               & 31.40               & 51.84               & 20.44                & 42.20               & 59.00                & 16.80                 & 18.62                         \\ \midrule
\multirow{2}{*}{Drink}                                                      & Hot Drinks                                                                & 14.93              & 36.29               & 21.36                & 43.48              & 56.52               & 13.04                & 17.20                          \\ \cmidrule{2-9}
                                                                            & Rice Drinks                                                               & 35.56              & 47.87               & 12.31                & 42.00               & 54.67               & 12.67                & 12.49                         \\ \midrule
\multirow{2}{*}{Book}                                                       & Computer Magazines                                                        & 12.26              & 50.95               & 38.69                & 23.33              & 64.00                & 40.67                & 39.68                         \\ \cmidrule{2-9}
                                                                            & Music Magazines                                                           & 43.11              & 51.69               & 8.58                 & 46.21              & 52.87               & 6.66                 & 7.62                          \\ \midrule
Drug                                                                        & Largest Selling Pharmaceutical Products                                   & 32.00               & 48.86               & 16.86                & 77.14              & 81.71               & 4.57                 & 10.72                         \\ \midrule
Sport                                                                       & National Basketball Association arenas                                    & 75.88              & 80.42               & 4.54                 & 88.46              & 88.46               & 0.00                  & 2.27                          \\ \midrule
GPE                                                                         & Capitals in China                                                         & 73.32              & 81.11               & 7.79                 & 84.85              & 84.85               & 0.00                  & 3.90                          \\ \midrule
Building                                                                    & Museums in China                                                          & 44.91              & 70.84               & 25.93                & 57.80               & 73.40                & 15.60                 & 20.77                         \\ \midrule
\multirow{3}{*}{Nature}                                                     & Nearest Exoplanets                                                        & 43.51              & 62.80                & 19.29                & 47.74              & 67.10                & 19.36                & 19.33                         \\ \cmidrule{2-9}
                                                                            & Submarine Volcanoes                                                       & 11.63              & 41.15               & 29.52                & 30.43              & 47.83               & 17.40                 & 23.46                         \\ \cmidrule{2-9}
                                                                            & World Heritage Sites in the United States                                 & 23.41              & 41.90                & 18.49                & 34.55              & 46.67               & 12.12                & 15.31                         \\ \midrule
Education                                                                   & Double First Class Universities in China                                  & 86.40               & 96.75               & 10.35                & 89.80               & 97.00                & 7.20                  & 8.78                          \\ \midrule
Internet                                                                    & Internet Forums                                                           & 17.99              & 40.12               & 22.13                & 32.73              & 45.45               & 12.72                & 17.43                         \\ \midrule
\multirow{2}{*}{Military}                                                   & United States Air Force Installations                                     & 71.86              & 86.36               & 14.50                 & 80.60               & 90.80                & 10.20                 & 12.35                         \\ \cmidrule{2-9}
                                                                            & World War II Prisoner-of-war Camps in the United States                   & 54.91              & 73.97               & 19.06                & 67.00               & 75.00                & 8.00                  & 13.53                         \\ \midrule
\multirow{2}{*}{Mathematics}                                                & Mathematical Topics in Classical Mechanics                                & 27.62              & 27.76               & 0.14                 & 48.72              & 46.15               & -2.57                & -1.22                         \\ \cmidrule{2-9}
                                                                            & Mathematical Topics in Quantum Theory                                     & 55.90               & 41.33               & -14.57               & 72.00               & 56.00                & -16.00                & -15.29                        \\ \midrule
\multirow{2}{*}{Medicine}                                                   & Cancer Types                                                              & 70.31              & 79.94               & 9.63                 & 76.80               & 82.60                & 5.80                  & 7.72                          \\ \cmidrule{2-9}
                                                                            & Feline Diseases                                                           & 14.04              & 34.51               & 20.47                & 26.67              & 48.89               & 22.22                & 21.35                         \\ \midrule
\multicolumn{2}{l}{\textbf{Overall}}                                                                                                                    & \textbf{40.73}     & \textbf{58.36}      & \textbf{17.63}       & \textbf{51.99}     & \textbf{65.04}      & \textbf{13.05}       & \textbf{15.34}                \\ \bottomrule
\end{tabular}}
\label{tab:each_class_map}
\end{table*}

\subsection{Parameter analysis}
\label{appendix:parameter}

\begin{figure}[th]
    \centering
    \includegraphics[width=0.98\columnwidth]{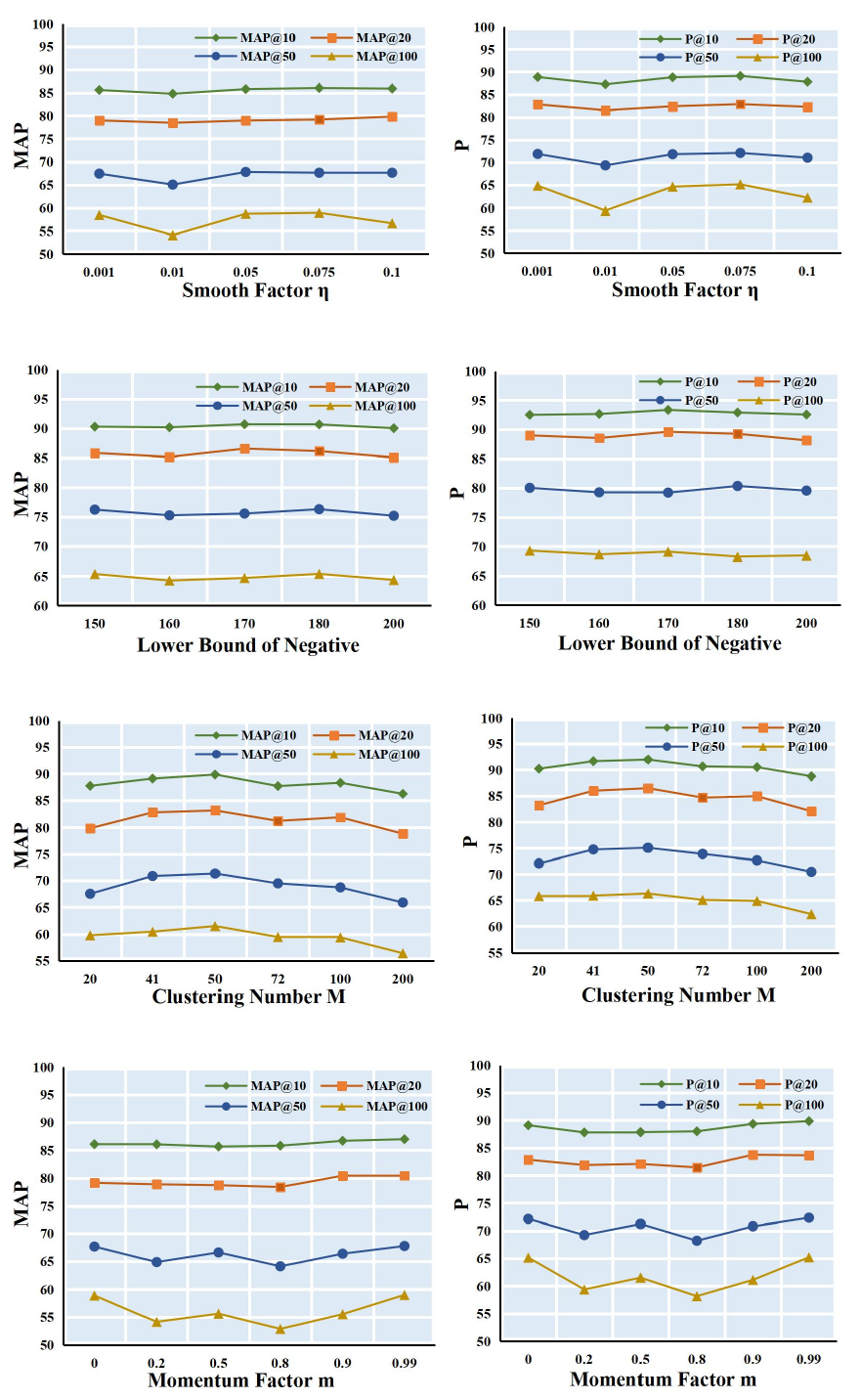}
    \caption{
    Sensitivity analysis of the hyperparameters in MultiExpan.
    }
    \label{fig:para}
\end{figure}

To investigate the sensitivity of MultiExpan with respect to its hyperparameters, we conducted an extensive analysis by varying the smoothing factor $eta$, the lower bound of negative $L_{neg}$, the clustering number $M$, and the momentum factor $m$. As depicted in Figure \ref{fig:para}, we observed the following findings:

(1) Both the smoothing and momentum factors aim at preventing MultiExpan from overfitting to the noise in the data, and their impact on the final performance of MultiExpan is relatively minor given the high quality of the dataset. Notably, the entity annotation provided by MESED is highly accurate, leading to a weak effect of the smoothing factor on MultiExpan.

(2) In contrastive learning, we found that the model performance is not significantly affected even if the value of lower bound of negative $L_{neg}$ is suboptimal. This is because the contrastive learning utilized by MultiExpan can effectively identify hard negative entities by assigning a higher loss penalty to negative samples that are more similar to positive samples in Equation \ref{eq:con3}.

(3) Regarding cluster learning, we assume that entities can be modeled by a probability distribution over $M$ target semantic classes. Our experimental results show that this assumption still holds true for $M$ larger than the number of ground truth clusters. In this case, the entities can be further represented as finer-grained semantic clusters. Consequently, we observed a modest increase in model performance when $M$ is slightly larger than the number of ground truth semantic classes, i.e., 41.

\subsection{Case study}
\label{append_sec_case}
\begin{figure*}
    \centering
    \includegraphics[width=1.98\columnwidth]{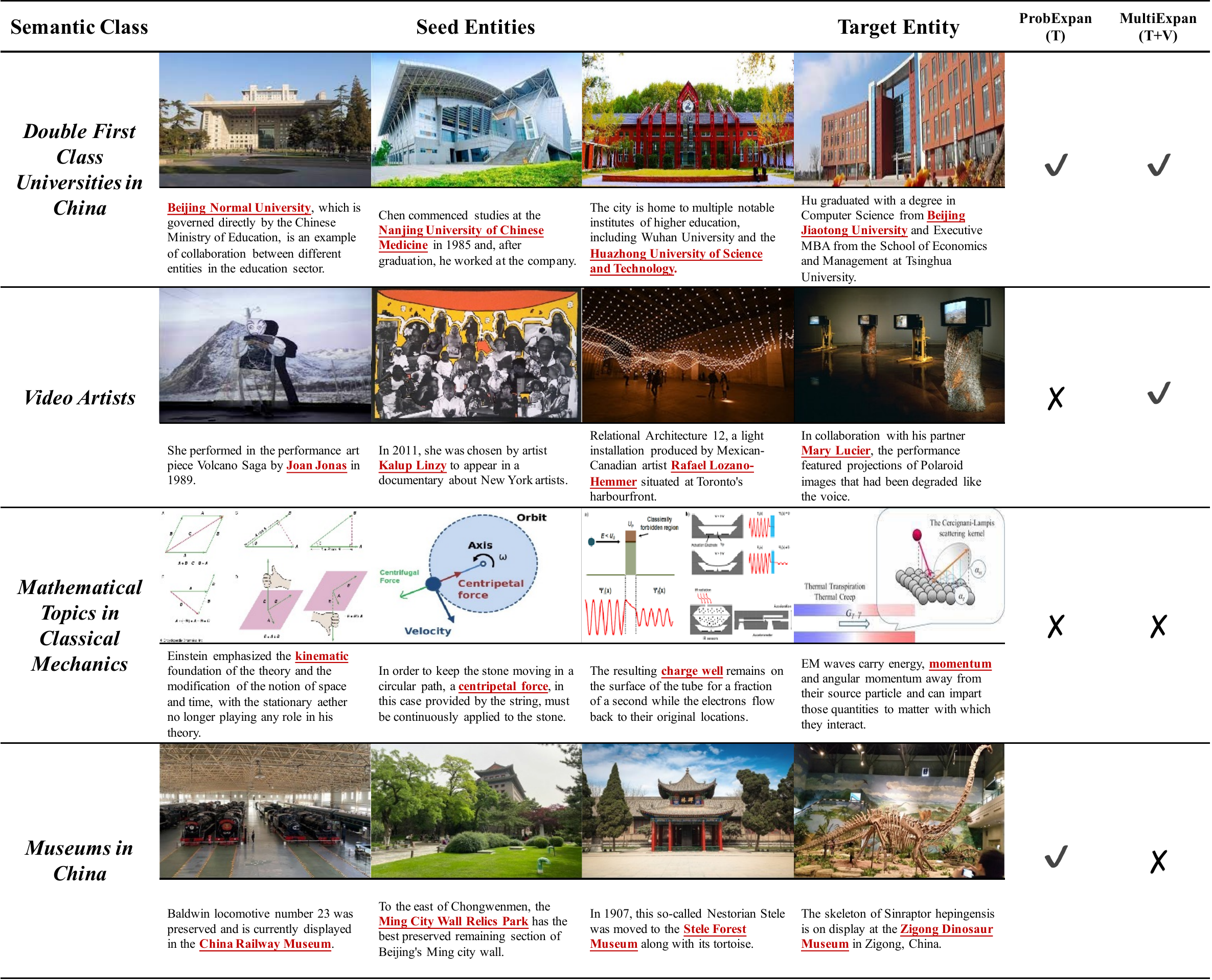}
    \caption{
    Case Studies. A \Checkmark mark means the target entity is in the top 100 of ranked entity list, a \XSolidBrush mark means the opposite.
    }
    \label{fig:case_study}
\end{figure*}
We present a series of intriguing case studies in Figure \ref{fig:case_study} to illustrate the utilization of multi-modal information. In the first row, due to the relative simplicity of the semantic class \texttt{Double First Class Universities in China}, textual information alone is sufficient to accurately expand the target entity \textit{Beijing Jiaotong University}. In contrast, the concept of \texttt{Video Artists} in the second row cannot be generalized by textual information alone in the ESE model. By displaying their artwork in a unified manner, the images offer an implicit visual clue to the semantic class of \texttt{Artists}. Conversely, for the last two cases, the inclusion of images did not yield positive benefits, indicating that further improvements are required in image utilization strategies. In the third row, a large number of mathematical symbols and geometric shapes are present in the images, yet the current MultiExpan does not fully comprehend and harness them. This inspires us to explore better image utilization methods in the future by comprehending the fine-grained semantics of the objects in the images (e.g., text in the images, regional features). In the last row, inclusion of the image even led to a negative effect; the representation of the target entity \textit{Zigong Dinosaur Museum}, as a dinosaur skeleton resulted in confusion and the model erroneously excluded this entity. However, this error, which arose due to multiple semantic misdirections, can evidently be circumvented as there are explicit words in the sentence indicating that this entity is a museum. This prompts us to investigate how cross-modal interactions can be leveraged to exclude noisy or misleading information in mono-modal forms in the future.

\subsection{Visual clues}
\begin{figure*}
    \centering
    \includegraphics[width=1.98\columnwidth]{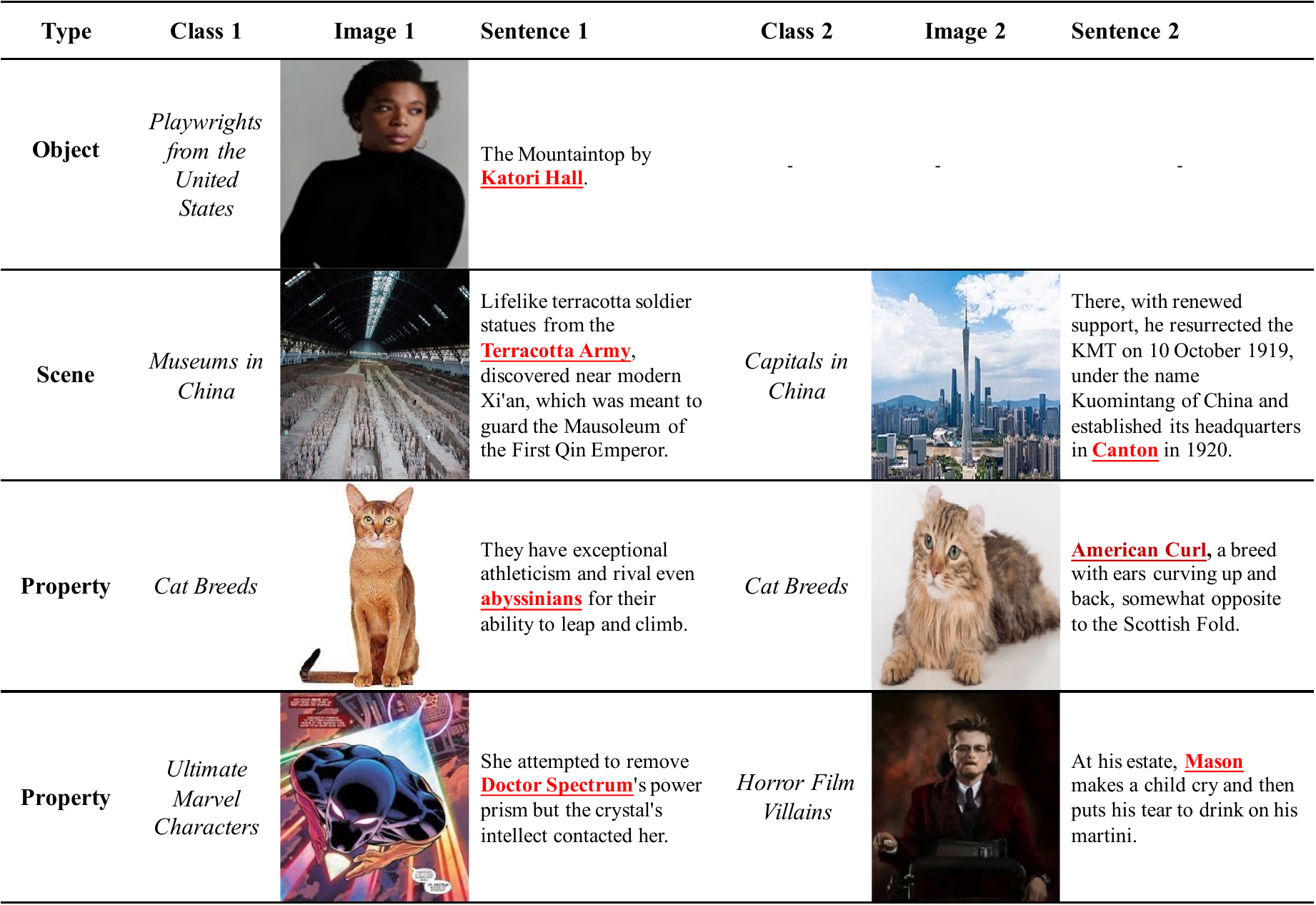}
    \caption{
    Examples of visual clues.
    }
    \label{fig:visual_clue}
\end{figure*}
In Figure \ref{fig:visual_clue}, we present some examples of visual clues, which contain objects, scenes and properties: (1) Objects, which can augment the limited textual information by depicting the entities themselves. For instance, the portrait featured in the first row not only confirms the entity as a person but also implies that the Mountaintop referred to in the brief sentence is not a physical peak but a book written by her. (2) Scenes, which showcase the environment where the entity exists to differentiate between the target semantic class and the hard negative semantic class, e.g., water vs. land. The example in the second row distinguishes between the semantic classes of \texttt{Museums in China
} and \texttt{Capitals in China}, which are both Chinese locations, through indoor and outdoor scenes. (3) Properties, which demonstrate the common traits of entities to align entities of the same class. As demonstrated in the third row, the fur and claws of the various cat breeds constitute a unified visual clue for the model. Additionally, the uniform comic book style of \texttt{Ultimate Marvel Characters} in the fourth row provides significant enhancements as it enables the distinction of realistic characters in the hard negative semantic class of \texttt{Horror Film Villains}.

\end{document}